%% file: Main.tex
\documentclass[letterpaper,times]{IONconf-v2}
\usepackage[]{graphicx}    % We use this package in this document
\newcommand{\ignore}[1]{}  % {} empty inside = %% comment

\usepackage{amsmath,amssymb,amsfonts}
\usepackage{bm}
\usepackage{algorithm2e}
\usepackage{float}
\usepackage[labelformat=simple]{subcaption}

\usepackage{mathtools}
\usepackage{siunitx}
\usepackage{multirow}
\usepackage{wasysym}
\usepackage{csquotes}
\usepackage{tabularray} 

\usepackage[hidelinks]{hyperref}

\input{macros}

\articletype{Regular papers}%

\received{}
\revised{}
\accepted{}
\doi{10.1109/xxx.xx.xxxx}
\journalname{NAVIGATION}
\journalvolume{-}
\journalnumber{-}

\title{Ionospheric and Plasmaspheric Delay Characterization for Lunar Terrestrial GNSS Receivers with Global Core Plasma Model}

\author[1]{Keidai Iiyama}

\author[1]{Grace Gao}

\authormark{IIYAMA \textsc{et al}}

\address[1]{Department of Aeronautics and Astronautics, Stanford University, California, United States}

\corres{Grace Gao, Department of Aeronautics and Astronautics, Stanford University, California, United States \\ \email{gracegao@stanford.edu}}

\begin{document}

% The Abstract. The * indicates a section excluded from numbering.
\input{sections/0_abstract.tex}


\maketitle

\input{sections/1_introduction}
\input{sections/2_method}
\input{sections/3_0_results}

\input{sections/4_conclusion}

\section*{acknowledgements}
This material is based upon work supported by The Nakajima Foundation.

\nocite{*}
\printbibliography[title=References]

\end{document}

%% file: macros.tex
\usepackage{amsthm}
\usepackage[english]{babel}

\newcolumntype{I}{!{\vrule width 1.5pt}}

\newcommand{\newtext}[1]{\textcolor{black}{#1}}

%% file: sections/0_abstract.tex
\abstract[Abstract]{
Recent advancements in lunar positioning, navigation, and timing (PNT) have demonstrated that terrestrial GNSS signals, including weak sidelobe transmissions, can be exploited for lunar spacecraft positioning and timing. 
While GNSS-based navigation at the Moon has been validated recently, unmodeled ionospheric and plasmaspheric delays remain a significant error source, particularly given the unique signal geometry and extended propagation paths.
This paper characterizes these delays using the Global Core Plasma Model (GCPM) and a custom low-cost ray-tracing algorithm that iteratively solves for bent signal paths.
We simulate first-, second-, and third-order group delays, as well as excess path length from ray bending, for GNSS signals received at both lunar orbit and the lunar south pole under varying solar and geomagnetic conditions.
Results show that mean group delays are typically on the order of 1 m, but can exceed 100 m for low-altitude ray paths during high solar activity, while bending delays are generally smaller but non-negligible for low-altitude ray paths.
We also quantify the influence of signal frequency, geomagnetic $K_p$ index, and solar R12 index.
These findings inform the design of robust positioning and timing algorithms that utilize terrestrial GNSS signals.
}

%% file: sections/1_introduction.tex
\section{Introduction}
\label{sec:introduction}
In the past decade, significant progress has been made toward using terrestrial Global Navigation Satellite System (GNSS) signals for positioning, navigation, and timing (PNT) at the Moon. Studies have long shown that GNSS signals can be tracked at lunar distances, despite being far outside their intended coverage. This concept relies on side-lobe signals that spill over beyond Earth, providing a weak but usable navigation source in lunar orbit. Several feasibility analyses \citep{Capuano2017, Delepaut2020} used simulations that combine high-sensitivity GNSS receivers and onboard orbit filters, indicating that continuous GNSS navigation from Earth to the Moon is achievable in principle.

Recent on-orbit demonstrations have validated GNSS use near the Moon. 
In February 2025, NASA and the Italian Space Agency’s Lunar GNSS Receiver Experiment (LuGRE) receiver on a commercial lander successfully acquired GPS and Galileo signals in lunar orbit, which is 401,000 km from the GNSS satellite~\citep{parker2022lugre, insidegnss2025lugre}. 
During a one-hour tracking period, LuGRE attained a navigation fix, achieving $\sim$1.5 km position accuracy and $\sim$2 m/s velocity accuracy in lunar orbit.
In parallel, the European Space Agency is developing the Lunar Pathfinder satellite, slated for launch around 2025–2026, which will be equipped with the NaviMoon receiver that will test GNSS-based navigation in lunar orbit~\citep{giordano2022pathfinder}. 

Two primary challenges limit terrestrial GNSS positioning and timing performance at the Moon: the extreme propagation distance produces very weak signals, and the resulting geometry leads to poor dilution of precision.
A range of techniques has been proposed to mitigate these limitations, including time-differenced carrier phase navigation~\citep{Iiyama2024TDCPNavigation}, inter-satellite or differential measurements~\citep{Cheung2024PNT, Iiyama2023Diff, Delepaut2024isl}, and fusion with complementary sensors~\citep{Vila2025fusion}.
All of these approaches depend critically on minimizing the user equivalent range error (UERE), making accurate modeling of signal delays essential for robust positioning under low-observability conditions.

Among potential error sources, ionospheric and plasmaspheric delays are particularly important.
Dual-frequency receivers can cancel most of the ionospheric contribution, but at the cost of increased receiver noise, which can be critical when tracking weak side-lobe signals.
Single-frequency terrestrial users typically rely on empirical models such as the Klobuchar model~\citep{Klobuchar1987} or Galileo’s NTCM-G~\citep{Hoque2019}, but these models are not directly applicable to lunar users.
Unlike a near-vertical path to the terrestrial users, signals received by lunar users traverse the ionosphere and plasmasphere horizontally, following long paths dominated by plasmaspheric propagation.
Accurate delay prediction, therefore, requires explicit modeling of electron density in the plasmasphere.

The plasmasphere, a region of cold, diffuse plasma extending from roughly 1,000 km to beyond 20,000 km altitude, has been the focus of extensive modeling over the past decade.
Although its electron density is lower than that of the ionosphere, it can contribute significantly to the total electron content (TEC) along high-altitude or nighttime GNSS paths~\citep{KLIMENKO2015}.
Several empirical models are available, including the Global Core Plasma Model (GCPM)\citep{Gallagher2000GCPM}, IRI-Plas\citep{Umut2017IRIPlas}, NeQuick~\citep{Antonio2012}, and the Neustrelitz Plasmasphere Model (NSPM)~\citep{Jakowski2018NSPM}.
GCPM provides a flexible, density-based description of the plasmasphere; IRI-Plas extends the International Reference Ionosphere to GNSS altitudes; NeQuick offers a time-dependent three-dimensional representation of the topside ionosphere and plasmasphere; and NSPM assimilates ground and satellite data for near–real-time applications.

Previous studies have investigated plasmaspheric delays for GNSS receivers in low-Earth and geosynchronous orbits.
For example, \citet{Peng2017} developed a global plasmaspheric electron content model using LEO GNSS measurements, and \citet{Matsumoto2023} applied first-order delay corrections from GCPM to improve precise orbit determination of a geostationary relay satellite.
To the best of our knowledge, however, no prior work has characterized plasmaspheric delays for terrestrial GNSS signals observed in lunar orbit.
Understanding how main-lobe and side-lobe signals are affected under different solar and geomagnetic conditions is critical for designing robust lunar PNT systems.

The objective of this paper is to characterize the plasmaspheric delay experienced by terrestrial GNSS signals as observed by receivers in lunar orbit and on the lunar surface.
We employ the Global Core Plasma Model to simulate first-, second-, and third-order ionospheric and plasmaspheric group delays, along with excess path delays caused by ray bending.
GCPM is selected for its ability to incorporate variable geomagnetic and solar activity conditions.

The main contributions of this work are as follows.
First, we propose a computationally efficient ray-tracing algorithm that iteratively determines the bent ray path using a Nelder–Mead algorithm~\citep{Nelder1965Simplex} without requiring gradient evaluations.
Second, we present detailed simulations of first-, second-, and third-order ionospheric and plasmaspheric delays, including bending delays, for GNSS signals received at both the lunar south pole and in lunar orbit.
Third, we analyze how these delays vary with signal frequency, geomagnetic activity ($K_p$ index), and solar activity (R12 index).
This paper expands upon our earlier conference publication~\citep{iiyama2025plasma} by incorporating updated simulations, including the ray-tracing example presented in Section~\ref{sec:raytracing_example}.

The paper is structured as follows. In Section \ref{sec:sim_method}, we describe the simulation method to characterize the plasmaspheric delay. In Section \ref{sec:sim_results}, we present the simulation results, and in Section \ref{sec:conclusion}, we conclude the paper and discuss future work.

%% file: sections/2_method.tex
\section{Simulation Method to Compute Ionospheric and Plasmaspheric Delays of GNSS Signals for Lunar Users}
\label{sec:sim_method}

\subsection{Ionosphere and Plasmasphere Simulation with GCPM}
The Global Core Plasma Model (GCPM)\citep{Gallagher2000GCPM} is an empirical electron–density model of the inner magnetosphere that provides both total electron density and the relative concentrations of major ions.
It integrates region-specific submodels for the ionosphere (based on the International Reference Ionosphere, IRI\citep{Bilitza2024IRI}), the plasmasphere, plasmapause, magnetospheric trough, and polar cap, with smooth transitions enforced by hyperbolic-tangent functions.
Electron density is specified as a function of the $K_p$ geomagnetic index, date (to set solar activity parameters for IRI), and position in solar magnetic (SM) coordinates, defined as follows~\citep{Matsumoto2023}:
\begin{itemize}
\item $X_{SM}$: X–Z plane contains the Sun and Earth.
\item $Y_{SM}$: Completes the right-handed system.
\item $Z_{SM}$: Aligned with Earth’s geomagnetic pole.
\end{itemize}

\begin{figure}[ht!]
    \centering
\includegraphics[width=0.9\columnwidth]{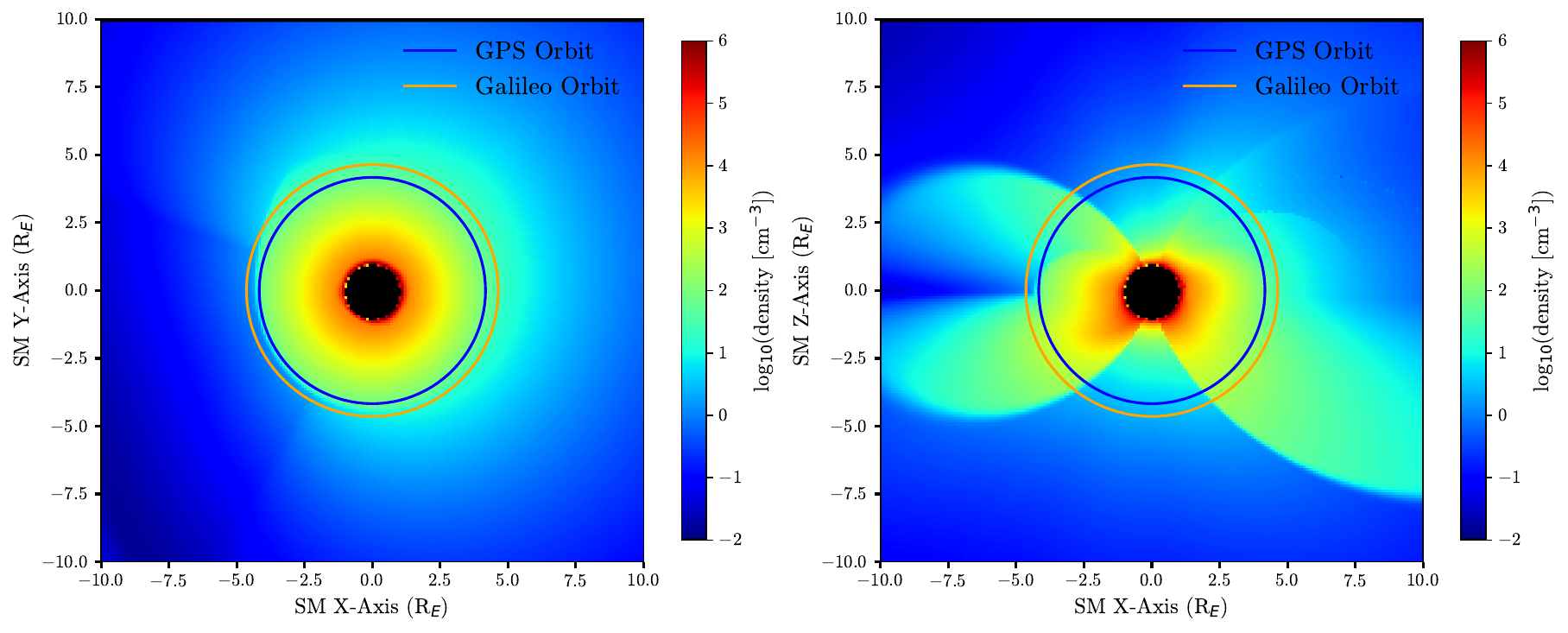}
\caption{Simulated plasmasphere electron density (log scale) for 1 Jan 2025, 12:00 UTC, $K_p=3.0$, using GCPM v2.4. Left: equatorial (X–Y) plane. Right: meridional (X–Z) plane in SM coordinates. Black region represents the Earth; blue and orange lines indicate GPS and Galileo satellite altitudes.}
\label{fig:gcpm_electron}
\end{figure}

Figure~\ref{fig:gcpm_electron} illustrates the electron density predicted by GCPM v2.4 for 1 January 2025, 12:00 UTC ($K_p=3.0$), showing the equatorial (X–Y) and meridional (X–Z) planes.

\subsection{Ionospheric and Plasmaspheric Delays}
For a right-hand circularly polarized signal, such as GNSS signals, the phase refractive index $n$ and the group refractive index $n_{gr}$ is given by ~\citep{Jakowski12}
\begin{align}
    n &= 1 - \frac{f_p^2}{2 f^2} - \frac{f_p^2 f_g \cos \theta}{2 f^3} - \frac{{f_p}^2}{4 f^4} \left[ \frac{{f_p}^2}{2} + {f_g}^2 (1 + \cos^2 \theta) \right] \\
    n_{gr} &= 1 + \frac{f_p^2}{2 f^2} + \frac{f_p^2 f_g \cos \theta}{f^3} + \frac{3{f_p}^2}{4 f^4} \left[ \frac{{f_p}^2}{2} + {f_g}^2 (1 + \cos^2 \theta) \right]
\end{align}
where $f_p = n_e e^2 / (4 \pi \epsilon_0 m)$ and $f_g = e B / (2 \pi m)$ are the plasma and gyro frequencies, respectively. Above, $e$ is the electron charge, $\epsilon_0$ is the permittivity of free space, $m$ is the electron mass, $B$ is the magnetic field strength, $f$ is the frequency of the signal, and $\theta$ is the angle between the ray path and the magnetic field line~\citep{Hoque2011BendingCorrection}.

The ionospheric and plasmaspheric phase advance $d_I$ and group delay $d_{Igr}$ of the GNSS signal along the path is given by
\begin{align}
    d_I &= \int (1 - n) ds = \frac{p}{f^2} + \frac{q}{2f^3} + \frac{u}{3f^4} \\
    d_{Igr} &= \int (n_{gr} - 1) ds = \frac{p}{f^2} + \frac{q}{f^3} + \frac{u}{f^4} \\
    p &= 40.3 \int n_e ds = 40.3 TEC = 40.3 (TEC_{LOS} + \Delta TEC_{bend})\\
    q &= -2.2566 \times 10^{12} \int n_e B \cos \theta ds \\
    u &= 2437 \int n_e^2 ds + 4.74 \times 10^{22} \int n_e B^2 (1 + \cos^2 \theta) ds
\end{align}
where $ds$ is the ray path arc, the quantities $d_{I1} = p/f^2$, $d_{I2}=q/f^3$, and $d_{I3} = u/f^4$ are the first-order, second-order, and third-order group delays, respectively. 
The integral of the electron density $n_e$ along the ray path is called the total electron content (TEC). 
Since the ray path is curved in the ionosphere and plasmasphere, TEC is split into the line-of-sight (LOS) TEC ($TEC_{LOS}$) and $\Delta TEC_{bend}$, which is the difference between the TEC along the curved ray path and the LOS between the transmitter and receiver.
The first-order delay is the dominant term, while the second- and third-order delays are below $1 \%$ of the first-order delays, for terrestrial users~\citep{Hoque2008Estimate} and low-earth orbit receivers~\citep{Hoque2011BendingCorrection}.

Ray bending also increases the physical path length, adding an excess path delay:
\begin{equation}
    d_{len} = \int ds - \rho
\end{equation}
where $\rho$ is the geometric distance between the transmitter and the receiver. 
The procedure for computing the ray path is described in the next section.

For the remainder of the paper, we focus on group delays.
The total group delay $d_{total}$ compared to the LOS distance is then given by the sum of the ionospheric and plasmaspheric group delay and the bending delay:
\begin{equation}
    d_{total} = d_{Igr} + d_{len} = d_{I1, LOS} + d_{I2} + d_{I3} + d_{I1, B} + d_{len}
\end{equation}
where
\begin{equation}
\begin{cases}
    d_{I1, LOS} = 40.3 \ TEC_{LOS} & \text{First Order LOS Group Delay} \\
    d_{I2} = \frac{q}{f^3}  &  \text{Second Order Group Delay} \\
    d_{I3} = \frac{u}{f^4} &  \text{Third Order Group Delay} \\
    d_{I1, B} = 40.3 \Delta TEC_{bend} & \text{Additional first order delays due to TEC contributions to ray-path bending}   \\
    d_{len} = \int ds - |r_{tx} - r_{rx}|  & \text{Excess path length due to bending}
\end{cases}
\end{equation}

\subsection{Ray Tracing Algorithm}
The propagation of the GNSS signal is governed by the vector eikonal equation~\citep{Born1980Optics}, as follows:
\begin{align}
    \frac{dt}{ds} &= \frac{n}{c} \\
    \frac{d \bm{r}}{ds} &= \hat{\bm{s}} \\
    \frac{d \bm{\hat{s}}}{ds} &= \newtext{\frac{\bm{\nabla n}  - \bm{\hat{s}} (\bm{\hat{s}} \cdot \bm{\nabla n})}{n} }
    \label{eq:ode}
\end{align}
where $s$ is the arc length, $t$ is the time, $\bm{r}$ is the position vector, $c$ is the speed of light, $\bm{\hat{s}}$ is the unit vector in the direction of propagation, and $n$ is the phase refractive index. 

In addition to the differential equations above, the trajectory of the ray is constrained by the boundary conditions at the transmitter and receiver~\citep{Caruso2023RayTracing}.
\begin{align}
    t (s = 0) &= t_{tx}  & \bm{r} (t_{tx}) &= \bm{r}_{tx} \\
    t (s = s_f) &= t_{rx}  & \bm{r} (t_{rx}) &= \bm{r}_{rx}
\end{align}
Note that given the reception time and position of the receiver, the time and position of the transmitter can be solved by iteratively solving the following light-time equation:
\begin{equation}
    t_{tx} = t_{rx} - \frac{||\bm{r}_{rx} - \bm{r}_{tx}(t_{tx})||}{c}
\end{equation}
where $\bm{r}_{tx}(t_{tx})$ is the position of the transmitter at time $t_{tx}$.
In order to compute the ray path, we need to solve for the unknown initial ray direction $\bm{\hat{s}_0}$ and the ray length $s_f$ that satisfies the boundary conditions. 
Several methods have been proposed to solve this problem, such as the shooting method (homing-in approach)~\citep{Jones1975RayTracing, Strangeways2000HomingIn} and the variational methods~\citep{Coleman2011RayTracing}.

\begin{figure}[ht!]
    \centering
    \includegraphics[width=0.75\textwidth]{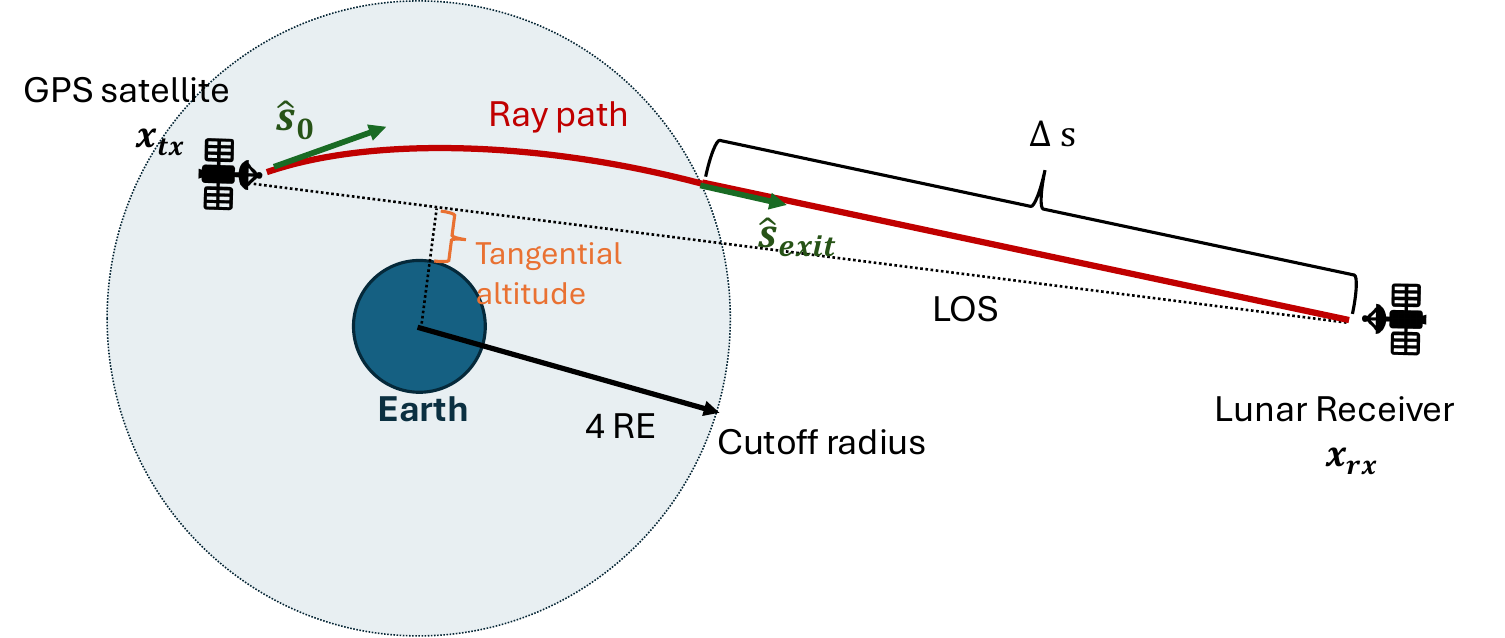}
    \caption{The ray tracing problem for the GNSS signals received by a lunar user. The ray path is divided into the plasmasphere section and the vacuum section, based on the observation that the electron density is significantly lower in higher altitude regions past the plasmapause. In this paper, we assume the ray path is straight once it exits the cutoff radius of 4 RE.}
    \label{fig:ray_tracing}
\end{figure}

Here, we solve the ray tracing problem using a shooting method accounting for the geometry of the raypath between the GPS satellite and the lunar receiver, as illustrated in Figure \ref{fig:ray_tracing}.

We first divide the ray path into the plasmasphere section and the vacuum section, based on the observation that the electron density is significantly lower in higher altitude regions above 2 to 3 Earth radii (RE)~\citep{Gallagher2000GCPM}. We assume that the ray path is straight once it exists beyond this boundary (set as 4 RE from Earth's center in this work). 
With this assumption, $s_f$ that minimizes the distance between the propagated ray and the receiver can be computed as follows:
\begin{equation}
    \begin{aligned}
        s_f &= s_{exit} + \Delta s \\
        \Delta s &= (\bm{x}_{rx} - \bm{x}_{exit}) \cdot \bm{\hat{s}_{exit}} 
        \label{eq:sf_exit}
    \end{aligned}
\end{equation}
where $s_{exit}, \bm{x}_{exit}, \bm{\hat{s}_{exit}}$ are the arc length, position and the unit vector of the ray at the exit point of the plasmasphere, which can be computed by integrating the ray tracing equations from the transmitter to the exit point.

Therefore, with this assumption, the ray tracing problem can be solved by searching for the initial ray direction $\bm{\hat{s}_0}$ that satisfies the boundary conditions. Since it is expensive to evaluate the electron density required for the evaluation of $d \bm{\hat{s}} / ds $, we run the following correction to correct for the initial direction $\bm{\hat{s}_0}$:

\begin{enumerate}
    \item Compute the ray path from the transmitter until it reaches the cutoff radius (4 RE) by integrating the ray tracing equations using the fourth-order Runge-Kutta method at discrete points $s_0, s_1, \ldots, s_{exit}$ along the ray. We compute the raypath by integrating Equation \eqref{eq:ode}.  We use an adaptive stepsize based on the altitude of the ray, where we use $\Delta s = 10$ km when the altitude is below 1000 km, $\Delta s = 20$ km when the altitude is below 4000 km, and $\Delta s = 100$ km otherwise. We store the $ d\bm {\hat{s}} / ds $ at each point.
    \item Terminate integration once the ray reaches the cutoff radius. Evaluate the final position $\bm{x}_{rx}$ of the ray at the receiver using the following equation:
    \begin{equation}
        \bar{\bm{x}}_{rx} = \bm{x}_{exit} + \Delta s \bm{\hat{s}_{exit}}
    \end{equation}
    \item  Run the Nelder-Mead optimization algorithm~\citep{Nelder1965Simplex} to compute the initial ray direction that minimizes the distance between the propagated final position $\bar{\bm{x}}_{rx} $ and the receiver position $\bm{x}_{rx}$. 
    It has been found that the Nelder-Mead simplex algorithm works reliably when solving for the initial ray direction $\bm{\hat{s}_0}$~\citep{Hoque2008Estimate}.
    Instead of directly optimizing for the direction vector $\bm{\hat{s}_0}$, we optimize for the azimuth and elevation angles of the initial ray direction, which turns into a 2-D optimization problem.
    In the propagation for the optimization, we use the stored $ d \bm{\hat{s}} / ds $ at each point to compute the ray path, instead of evaluating the electron density at each point, to avoid calling the GCPM at each iteration.
    \item Repeat 1-3 until the distance between the propagated final position $\bar{\bm{x}}_{rx} $ and the receiver position $\bm{x}_{rx} $ falls below a specified threshold, or until the terminal position error stops decreasing.
\end{enumerate}

Without any correction, the final position error is on the order of 10--100 km, due to the bending of the ray path in the plasmasphere. 
By applying the correction, the difference between the propagated final position $\bar{\bm{x}}_{rx}$ and the receiver position $\bm{x}_{rx}$ can be reduced to within sub-meter to 100 m, ensuring convergence of the computed total ionospheric and plasmaspheric delay as described in the next section.

\subsection{Ray Tracing Example}
\label{sec:raytracing_example}
To illustrate the performance of the proposed raytracing algorithm, in this section, we show an example raytracing result between a transmitter (PRN 4 GPS satellite) located at $\mathbf{r}_{tx} = \begin{bmatrix} 24513.42 & 1876.09 & 10266.99 \end{bmatrix}$ km and a lunar receiver located at $\mathbf{r}_{rx} = \begin{bmatrix} -343532.59 & -125200.76 & -123527.20 \end{bmatrix}$ km in the J2000 frame. 
The tangential altitude of the ray is 163.5 km, passing through the ionosphere.

\begin{figure}[ht!]
\begin{minipage}{0.47\linewidth}
    \centering
    \includegraphics[height=70mm]{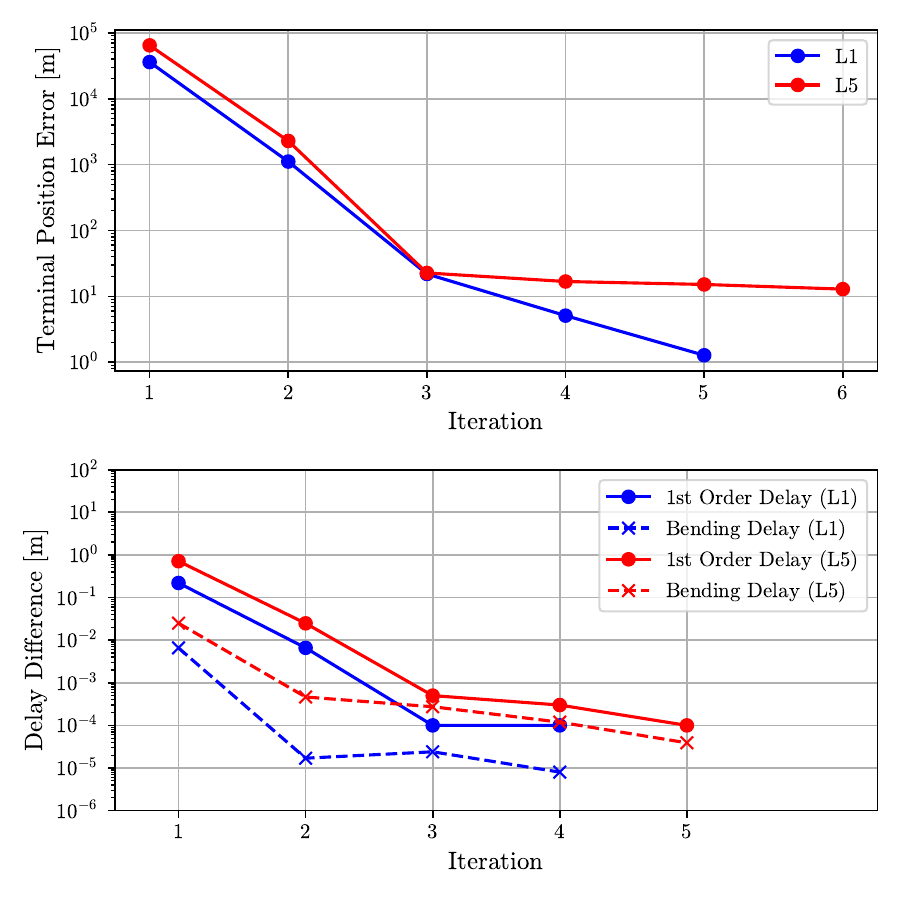}
    \caption{The terminal position error (top) and the differences to the converged delay value (bottom) over iterations. The L1 signal converges faster than the L5 signals, which experience larger delays. Three iterations lead to a delay simulator error below 1 mm. }
    \label{fig:convergence}
\end{minipage}
\hfill
\begin{minipage}{0.50\linewidth}
    \centering
    \includegraphics[height=72mm]{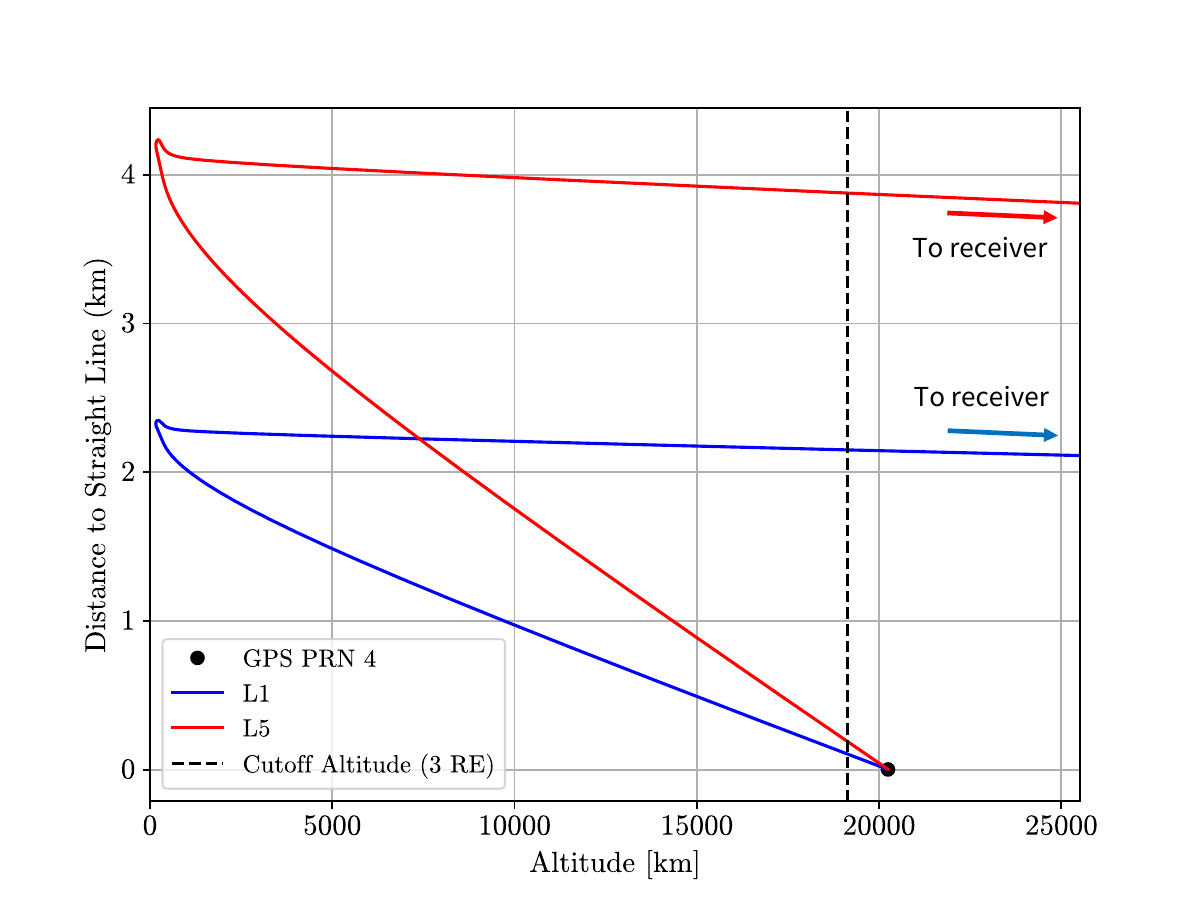}
    \caption{Distance between the ray path and the straight line connecting the transmitter and the receiver. The distance to the straight line reaches a maximum near the lowest altitude point. The L5 signal experiences larger bending than the L1 signal.}
    \label{fig:distance_line}
\end{minipage}
\hfill
\end{figure}

Figure~\ref{fig:convergence} shows the reduction of terminal position error and group-delay difference over iterations.
The L1 signal converged to a terminal error of 1.27 m in five iterations, whereas the lower-frequency L5 signal converged to 12.89 m in six iterations.
First-order and bending-delay errors dropped below 1 mm after three iterations, when the terminal position error was already $\sim$22 m.
Because the L1 signal experiences smaller bending, it converges faster and with smaller final position error.

Figure~\ref{fig:distance_line} plots the displacement between the bent ray path and the straight transmitter–receiver line.
The largest offset occurs near the point of lowest altitude, with the L5 signal exhibiting greater bending due to its lower frequency and correspondingly smaller refractive index.

%% file: sections/3_0_results.tex
\section{Simulation Results}
\label{sec:sim_results}

% Modeling with GCPM
% Characterize the distribution of the delay
\input{sections/3_1_simconfig.tex}
\input{sections/3_2_delaysim.tex}

\input{sections/3_3_frequency.tex}

%% file: sections/3_1_simconfig.tex
\subsection{Simulation Configuration}
\label{sec:sim_config}
\subsubsection{Lunar Satellites}
We simulated five users in the lunar orbit, which are the LCRNS satellites, with the initial orbital elements as shown in Table~\ref{tab:elfo_oe}.
The initial states were taken from the reference constellation~\citep{LCRNS2025}, and the initial epoch was set to March 1st, July 2027.
The propagated LCRNS satellite orbits are shown in Figure~\ref{fig:lcrns_orbits}.
We also simulated the GPS tracking for a user on the lunar south-pole, resulting in a total of six users in the simulation.

\subsubsection{GNSS Satellites and Tracking Simulations}
\begin{figure}[ht!]
\begin{minipage}[b]{0.48\textwidth}
    \centering
    \includegraphics[trim={0mm 0mm 0mm 5mm}, clip, width=0.95\textwidth]{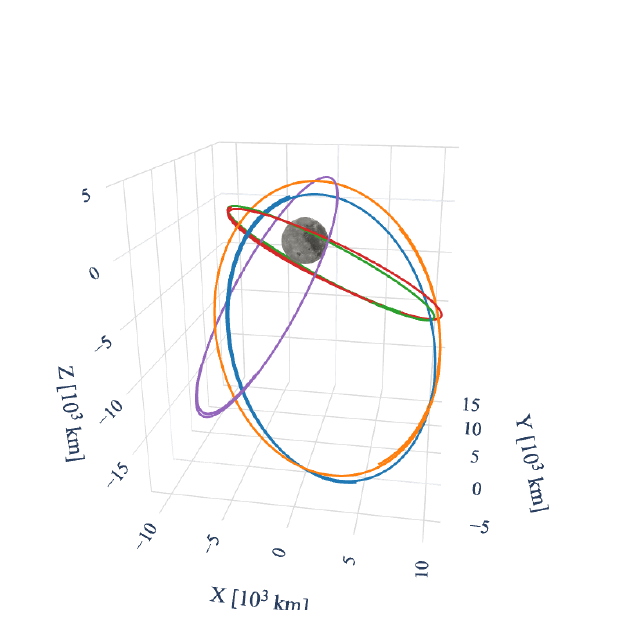}
    \caption{The orbits of the 5 LCRNS satellites in Moon-centered inertial frame}
    \label{fig:lcrns_orbits}
\end{minipage}
\hfill
\begin{minipage}[b]{0.48\textwidth}
    \centering
    \includegraphics[trim={0mm 0mm 0mm 5mm}, width=0.95\textwidth]{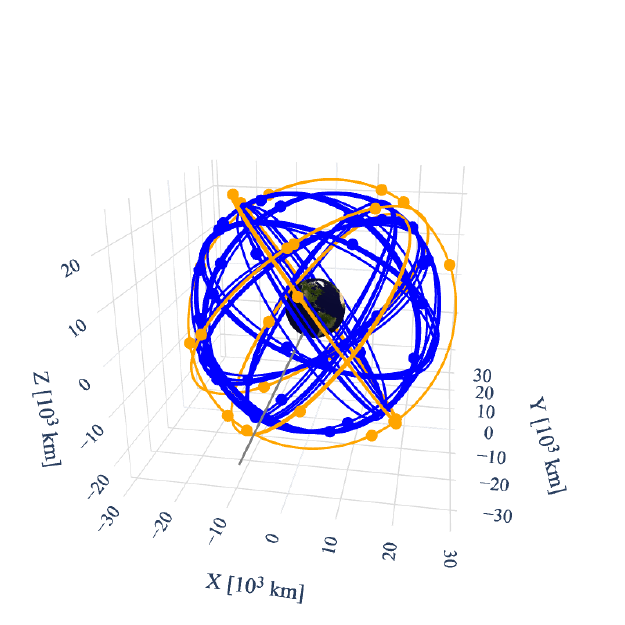}
    \caption{The GPS (blue) and GALILEO (orange) satellite constellation with the lunar direction for the initial epoch (2027/03/01 12:00:00 UTC)}
    \label{fig:gnss_constellation}
\end{minipage}
\end{figure}

The orbits of the GPS and GALILEO satellites were initialized using the SP3 files provided for the epoch June 23, 2025. The GPS and GALILEO satellite constellation with the lunar direction for the initial epoch is shown in Figure~\ref{fig:gnss_constellation}. 

\begin{figure}[ht!]
\centering
    \centering
    \includegraphics[width=0.95\textwidth]{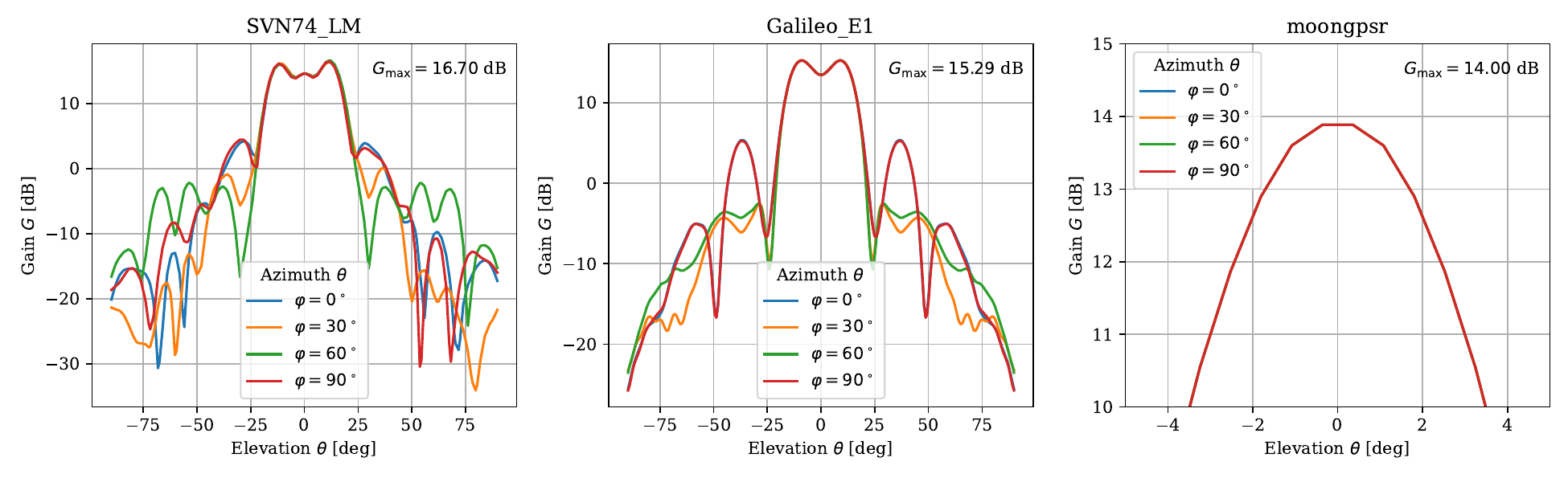}
    \caption{The antenna patterns of the selected GPS and GALILEO satellites and the lunar GNSS receiver}
    \label{fig:antenna_pattern}
\end{figure}

The GNSS antenna patterns were initialized using the references available online.
We used the parameters from the NASA antenna characterization experiment~(ACE) study~\citep{donaldson2020characterization} for Block II-F,  the original data by Lockheed Martin for Block IIR and IIR-M satellites \citep{Marquis2015}, the United States Coast Guard Navigation Center~\citep{Fischer2022} for Block-III, and Publications Office of the European Union~\citep{Menzione2024} for GALILEO satellites.
The antenna patterns of the selected GPS and GALILEO satellites and the GNSS receiver we assumed for the lunar users are shown in Figure \ref{fig:antenna_pattern}.

We assumed that the GNSS receiver of the lunar users have a peak gain of 14 dBi, with a half-power beamwidth of 6 degrees. We assumed that the GNSS receiver is able to track the signals from the GPS and GALILEO satellites with a C/N0 of 18 dB-Hz or higher, based on the experimental results of the NaviMoon receiver~\citep{NavimoonIOD}. The receiver noise errors of the pseudorange measurements are simulated assuming a thermal noise in the Delay Lock Loop (DLL), which depends on the carrier-to-noise ratio of the received signal~\citep{kaplan2017understanding}.

\begin{table}[htb]
  \caption{Initial Condition of the LCRNS Satellites}
  \label{tab:elfo_oe}
  \begin{tblr}{
    colspec={Q[l, 5.0cm]Q[c,1.0cm]X[c]X[c]X[c]X[c]X[c]},
    width=\textwidth,
    row{even} = {white,font=\small},
    row{odd}  = {bg=black!10,font=\small},
    row{1}    = {bg=black!20,font=\bfseries\small},
    hline{Z}  = {1pt,solid,black!60},
    rowsep=3pt
  }Parameter & Unit & LCRNS 1 & LCRNS 2 & LCRNS 3 & LCRNS 4 & LCRNS 5 \\
Semi-major axis $a$ & km  & 11315.4 & 11317.9 & 11305.4 & 11326.3 & 11307.9 \\
Eccentricity $e$    & –   & 0.69182 & 0.69182 & 0.69182 & 0.69182 & 0.69182 \\
Inclination $i$     & deg & 59.373  & 59.951732 & 52.733096 & 52.513419 & 56.310396 \\
Right ascension of ascending node $\Omega$ & deg & 321.019197 & 320.997768 & 81.148790 & 81.138818 & 204.889626 \\
Argument of periapsis $w$ & deg & 92.494031 & 92.505016 & 92.062891 & 92.068945 & 85.444071 \\
Mean anomaly $M_0$  & deg & 0 & 180.0 & 140.049207 & 195.992393 & 164.007607 \\
  \end{tblr}
\end{table}

\subsubsection{Software}
For simulation, we created a simulation software that interfaces the GCPM v2.4 model (implemented in Fortran) with the open-source lunar PNT library, LuPNT~\citep{Iiyama_LuPNT, Vila2025LuPNT}, which is developed in Python and C++. 
Several modifications were introduced to the original GCPM code to compute electron density values for recent epochs (from 2015 onward) and accept custom solar activity parameters (R12 index) as inputs.
The simulation software computes the GNSS and lunar satellite orbits using LuPNT and performs ray tracing by using the GCPM model to compute electron density values along the ray path.
We plan to release the simulation software as an open-source package.

%% file: sections/3_2_delaysim.tex
\subsection{Ionospheric and Plasmaspheric Delay Simulation Results}
\label{sec:delay_sim}

\subsubsection{Results for the Baseline Test Case}
\label{sec:sim_result_baseline}
The baseline scenario was simulated using electron densities for 1 January 2025, 12:00 UTC, with a simulation span of 45 hours—approximately \newtext{one and a half orbital periods} of the LCRNS satellites.
This epoch was selected because the GCPM model cannot be run for the 2027/03/01 target date due to unavailable historical parameters.
To represent moderate geomagnetic activity, we set the $K_p$ index to 3.0 (the actual value at the epoch was 6.667).
For solar activity, we used historical measurements of R12 = 167.24 and F10.7 = 210.39, corresponding to high solar activity expected near the peak of Solar Cycle 25.
As discussed in Sections~\ref{sec:geomagnetic} and~\ref{sec:solar_activity}, the resulting delays are sensitive to both geomagnetic and solar activity levels.

\begin{figure}[ht!]
    \centering
    \includegraphics[width=0.95\textwidth]{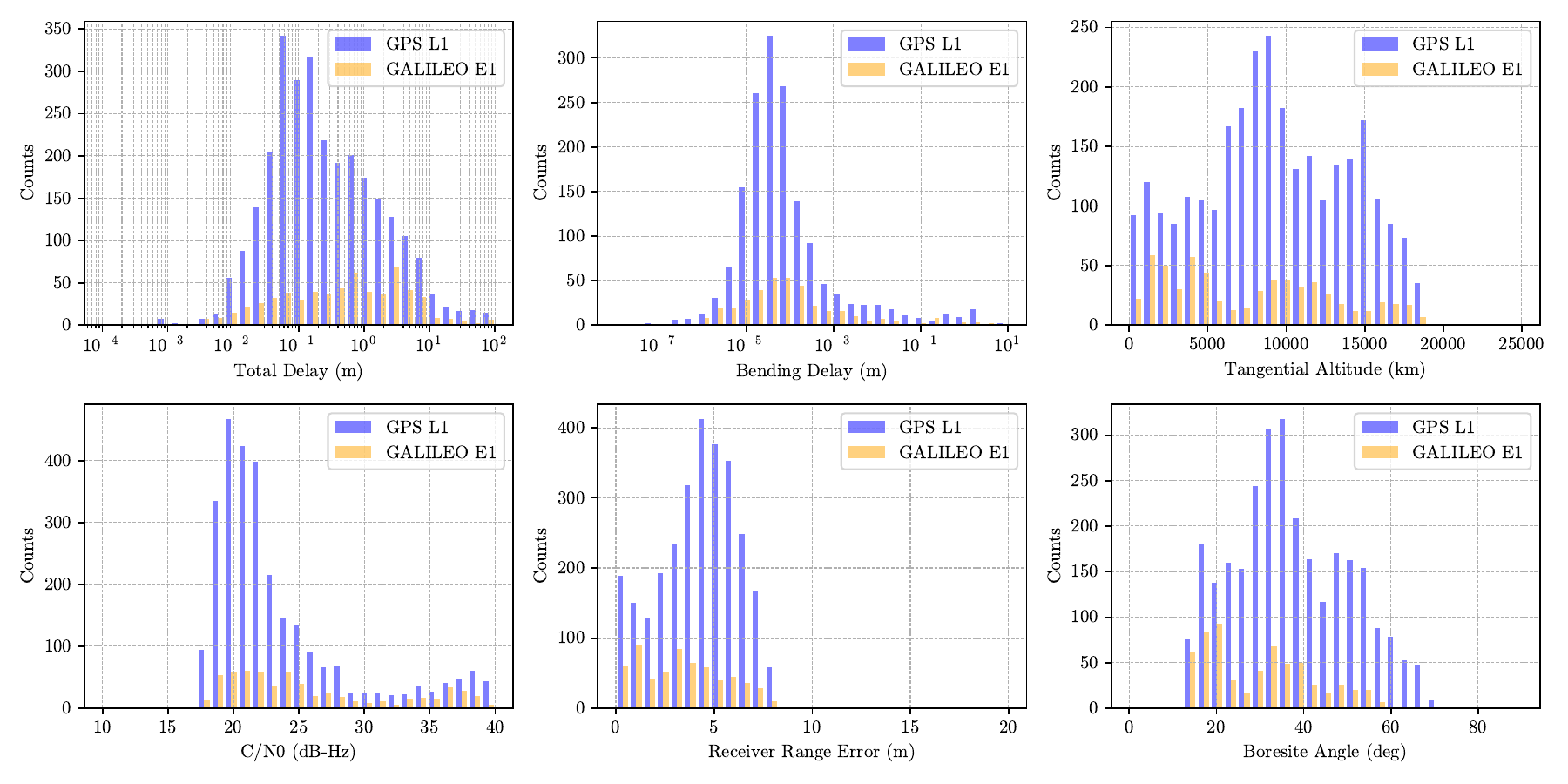}
    \caption{The histogram of the ionospheric and plasmaspheric delays and related parameters for the baseline scenario ($K_p = 3.0$, R12 = 167.24, 2025/01/01 12:00:00 UTC)}
    \label{fig:histogram_nominal}
\end{figure}

\begin{figure}[ht!]
    \centering
    \includegraphics[width=0.95\textwidth]{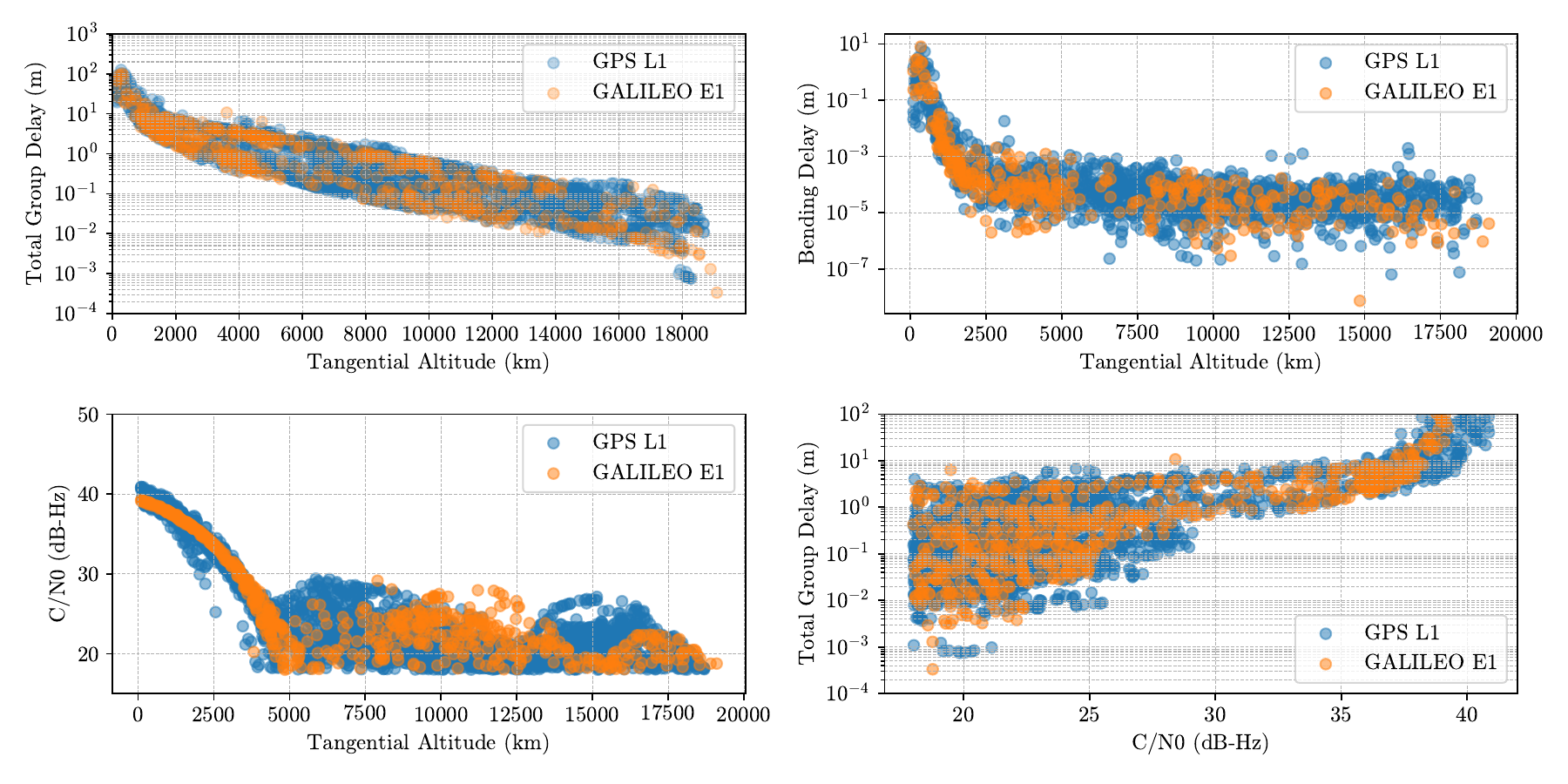}
    \caption{The scatter plot of the ionospheric and plasmaspheric delays for the nominal scenario. Both the group delay and the $C/N_0$ decrease as the tangential altitude increases, creating a trade-off between $C/N_0$ and group delay. ($K_p = 3.0$, R12 = 167.24, 2025/01/01 12:00:00 UTC)}
    \label{fig:scatter_nominal}
\end{figure}

Simulation outputs for GPS L1 and Galileo E1 signals are summarized in Table~\ref{tab:nominal_result} and illustrated in Figures~\ref{fig:histogram_nominal} and \ref{fig:scatter_nominal}.
The results include all signals received by the LCRNS satellites and a lunar south-pole user, sampled every 30 minutes along the orbits.

The analysis shows that ionospheric and plasmaspheric delays become significant for low-altitude ray paths.
For tangential altitudes below 500 km, the mean first-order group delay reaches up to 50.59 m, but the delay decreases rapidly with altitude: above 10,000 km, the mean group delay falls below 0.2 m.
Bending delays are much smaller than the first-order term but can still exceed 1 m for paths with tangential altitudes under 1000 km.
Notably, the extra delay caused by bending exceeds the actual increase in geometric path length.
Second- and third-order delays remain negligible, several orders of magnitude smaller than the first-order contribution.

The carrier-to-noise ratio ($C/N_0$) values reported in Table~\ref{tab:nominal_result} further highlight a bias–variance trade-off.
Signals with low tangential altitudes experience larger ionospheric and plasmaspheric delays (higher bias) but benefit from higher $C/N_0$ and thus lower receiver thermal noise (lower variance).
Conversely, high-altitude signals exhibit smaller propagation delays but lower $C/N_0$, resulting in higher receiver noise.
This interplay between delay and noise is a key consideration when using terrestrial GNSS signals in lunar orbits.

\begin{table}[htb]
  \caption{The Mean Group Delays and C/N0 values Observed at the Lunar Receivers for the Baseline Scenario ($K_p = 3.0$, R12 = 167.24, 2025/01/01 12:00:00 UTC. See Figure \ref{fig:ray_tracing} for the definition of the tangential altitude)}
  \label{tab:nominal_result}
  \begin{tblr}{
    colspec={X[c]X[c]X[c]X[c]X[c]X[c]X[c]X[c]},
    width=\textwidth,
    row{even} = {white,font=\small},
    row{odd}  = {bg=black!10,font=\small},
    row{1}    = {bg=black!20,font=\bfseries\small},
    row{2}    = {bg=black!20,font=\bfseries\small},
    hline{Z}  = {1pt,solid,black!60},
    rowsep=3pt
  }
Tangential Altitude [km] &
Total Delay [m] &
1st Order Delay (LOS) [m] &
2nd Order Delay [m] &
3rd Order Delay [m] &
Bending Excess Path [m] &
Bending Excess TEC [m]  &
Carrier-to-noise ratio [dB-Hz] \\
$h_{\perp} $ & $d_{total}$& $d_{I1,\mathrm{LOS}}$ & $d_{I2}$ & $d_{I3}$ & $d_{len}$ & $d_{I1,B}$ & $C/N_0$ \\
0–500     & 50.59 & 49.52 & -1.35e-02 & 9.96e-04 & 3.64e-01 & 7.18e-01 & 39.55 \\
500–1000  & 14.35 & 14.23 & -2.16e-03 & 8.34e-05 & 3.95e-02 & 7.93e-02 & 38.70 \\
1000–2000 & 6.28  & 6.28  & -3.69e-04 & 8.45e-06 & 8.04e-04 & 1.51e-03 & 36.50 \\
2000–3000 & 3.15  & 3.15  & 5.55e-05  & 1.61e-06 & 4.51e-05 & 1.14e-04 & 32.75 \\
3000–4000 & 2.41  & 2.41  & 8.46e-05  & 6.65e-07 & 1.14e-05 & 1.65e-04 & 27.64 \\
4000–6000 & 1.65  & 1.65  & -2.91e-05 & 4.04e-07 & 6.92e-06 & -4.65e-05 & 23.41 \\
6000–8000 & 0.63  & 0.63  & -2.50e-06 & 5.38e-08 & 8.04e-07 & -3.26e-06 & 23.07 \\
8000–10000& 0.35  & 0.35  & 5.92e-07  & 1.59e-08 & 2.49e-07 & 2.51e-06 & 21.30 \\
10000–15000&0.12  & 0.12  & -9.46e-08 & 1.96e-09 & 2.11e-08 & -1.21e-05 & 20.48 \\
15000–20000&0.04  & 0.04  & 2.06e-07  & 2.73e-10 & 1.20e-07 & 3.30e-06 & 20.94 \\
  \end{tblr}
\end{table}

\subsubsection{Effect of the Geomagnetic Activity ($K_p$ Index)}
\label{sec:geomagnetic}
\begin{figure}[ht!]
    \centering
    \includegraphics[width=0.95\textwidth]{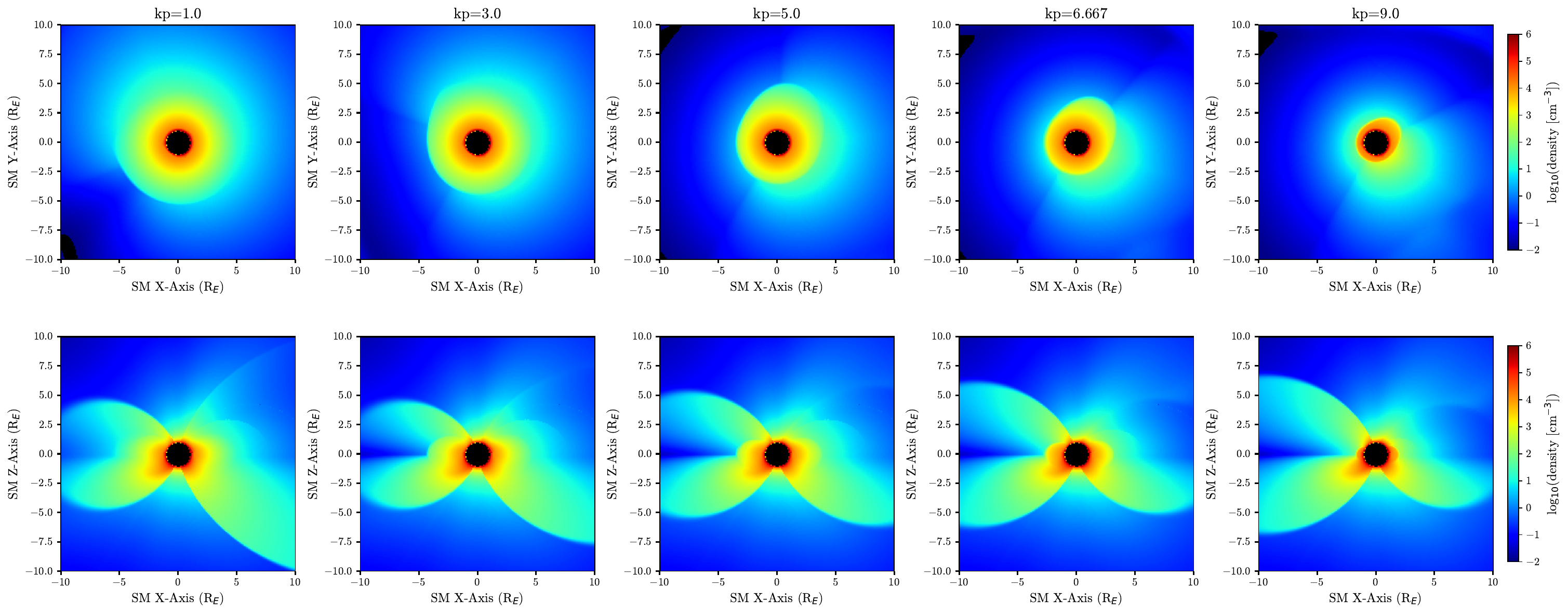}
    \caption{The electron density at the X-Y (top row) and X-Z (bottom row) plane for different $K_p$ indices. The epoch is set to January 1, 2025, 12:00:00 UTC, and R12 is set to 167.24 (value at the epoch). The plasmapause shrinks as the $K_p$ index increases, due to the erosion of the plasmasphere. }
    \label{fig:kp_ne_plots}
\end{figure}

To analyze the effect of geomagnetic activity on ionospheric and plasmaspheric delays, we simulated the electron density for five different $K_p$ indices, $K_p = 1, 3, 5, 6.667$, and $ 9.0 $. The electron density at the X-Y (equatorial) and X-Z (meridian) planes for different $K_p$ indices is shown in Figure~\ref{fig:kp_ne_plots}.
We observe that as the $K_p$ index gets higher, the plasmapause (the sharp boundary between the dense plasmasphere and the lower-density plasmatrough outside it) contracts from 4-6 RE at quiet conditions to 2-3 RE under strong geomagnetic activity, reflecting the erosion of the plasmasphere. 

\begin{figure}[ht!]
    \centering
    \includegraphics[width=0.95\textwidth]{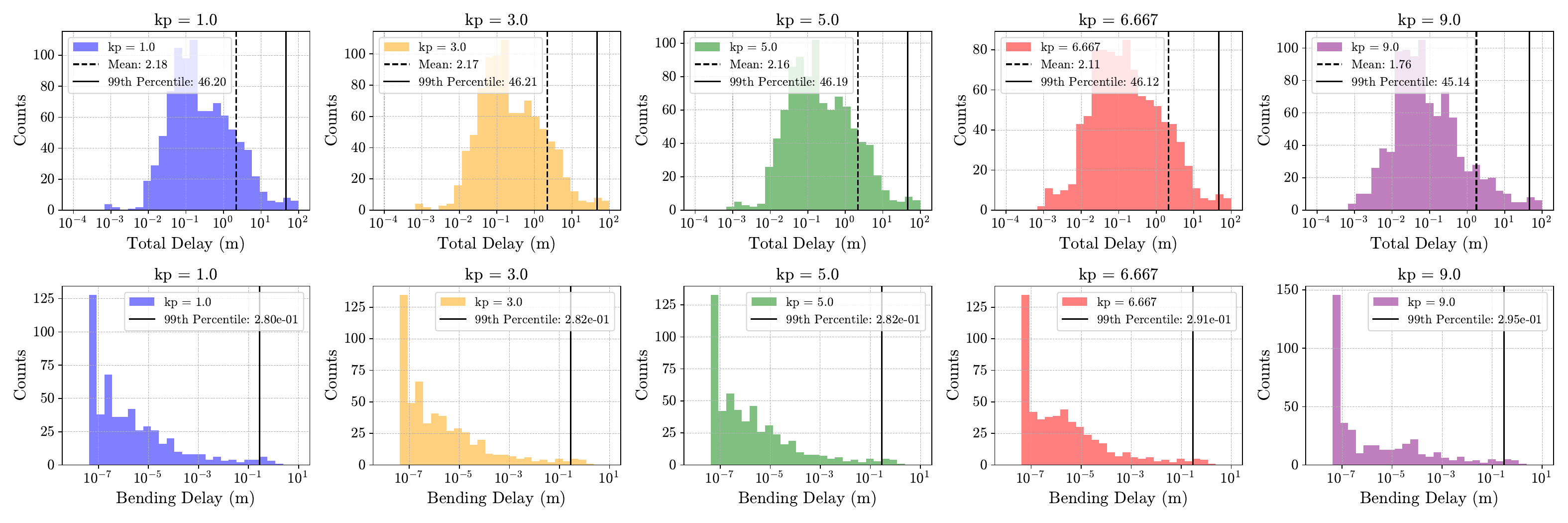}
    \caption{The histogram of the ionospheric and plasmaspheric delays for different $K_p$ indices. The delay becomes lower as the $K_p$ index increases, due to the shrinkage of the plasmapause.}
    \label{fig:histogram_kp}
\end{figure}

Figure~\ref{fig:histogram_kp} presents histograms of the ionospheric and plasmaspheric delays observed by LCRNS-1 and the lunar south-pole user across the different $K_p$ levels.
Consistent with the plasmapause contraction, higher $K_p$ indices produce generally smaller group delays, as the region of high electron density is reduced.

%%%%%%%%%%%%%%%%%%%%%%%%%%%%%%%%%%%%%%%%%%%%%%%%%%%%%%%%%%%%%%%
\subsubsection{Effect of Solar Activity (R12 Index)}
\label{sec:solar_activity}

To assess the influence of solar activity on ionospheric and plasmaspheric delays, we simulated electron densities for five levels of the R12 index:
$R12 = 10, 50, 100, 150,$ and $200$.
R12 is the 12-month smoothed sunspot number and serves as a standard indicator of solar activity.
In the GCPM model, the IRI submodel requires the F10.7 solar flux and IG12 ionospheric indices, which we derived from R12 using the empirical PHaRLAP relationships~\citep{pharlap}:
\begin{align}
\text{F10.7} &= 63.75 + 0.728 \times \text{R12} + 0.00089 \times \text{R12}^2, \\
\text{IG12} &= -12.349154 + 1.4683266 \times \text{R12} - 0.00267690893 \times \text{R12}^2.
\end{align}

\begin{figure}[ht!]
    \centering
    \includegraphics[width=0.95\textwidth]{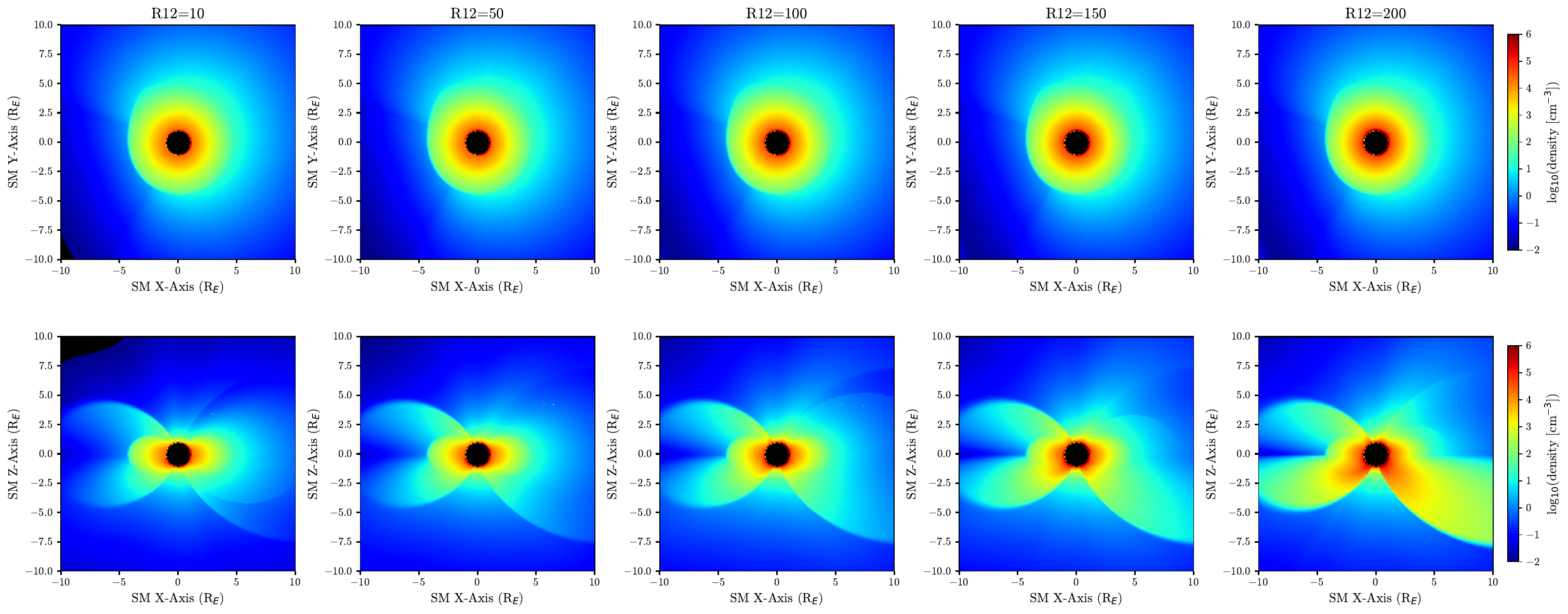}
    \caption{The electron density at the X-Y (top row) and X-Z (bottom row) plane for different R12 indices. The epoch is set to January 1, 2025, 12:00:00 UTC, and the $K_p$ index is set to 3 for all cases. The electron density is shown on a log scale.}
    \label{fig:r12_ne_plots}
\end{figure}

\begin{figure}[ht!]
    \centering
    \includegraphics[width=0.95\textwidth]{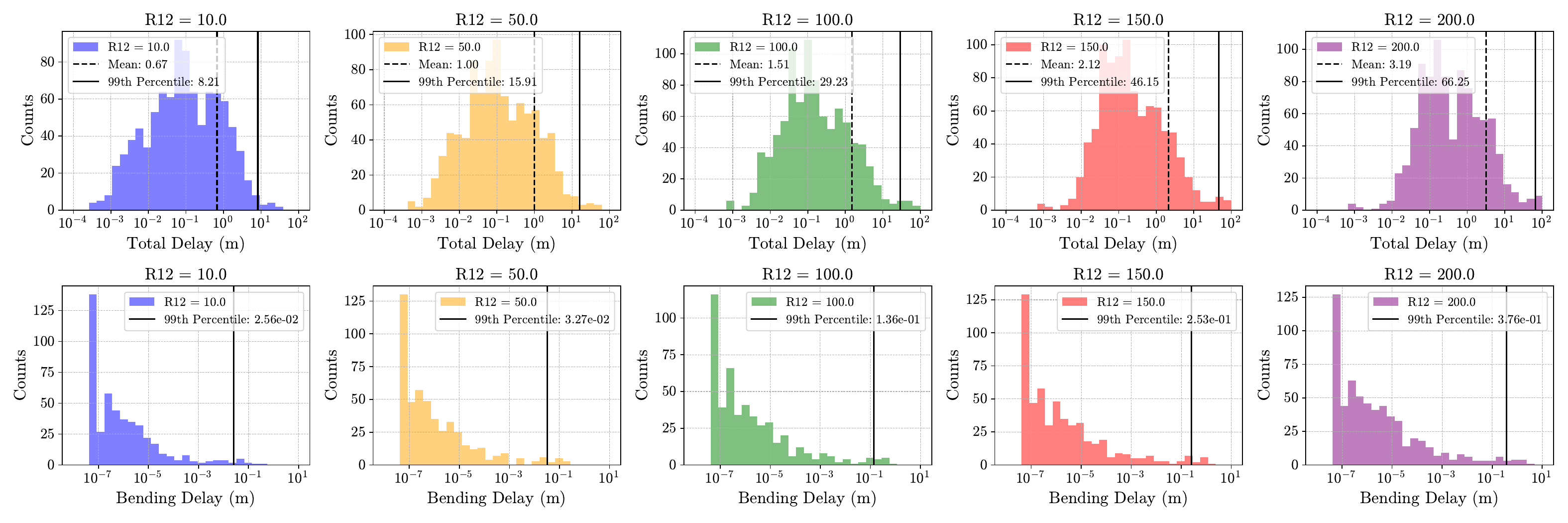}
    \caption{The histogram of the ionospheric and plasmaspheric delays for different R12 indices. The delay becomes larger as the R12 index increases, due to the increase in electron density in the ionosphere and plasmasphere.}
    \label{fig:histogram_r12}
\end{figure}

Electron density distributions for each R12 level in the X–Y (equatorial) and X–Z (meridional) planes are shown in Figure~\ref{fig:r12_ne_plots}.
The corresponding histograms of ionospheric and plasmaspheric delays observed by LCRNS-1 and the lunar south-pole user appear in Figure~\ref{fig:histogram_r12}.
As expected, higher R12 values produce larger group delays, reflecting the increase in ionospheric and plasmaspheric electron density during periods of elevated solar activity.

Table~\ref{tab:r12_result} summarizes the mean and 99th-percentile group delays across R12 levels and tangential altitudes.
These results provide practical guidance for setting altitude masks to exclude signals likely to contain extensive group delays.
For instance, to keep the 99$\%$ group delay below 5 m, an altitude mask of approximately 2000 km suffices when $R12 \le 50$, but the threshold must increase to 3000 km for $100 \le R12 \le 150$ and to 10,000 km when $R12 = 200$.

\begin{table}[ht!]
  \caption{The Mean and 99th Percentile Total Group Delays for Different R12 values and Tangential Altitudes (L1 signal, $K_p = 3.0$, 2025/01/01 12:00:00 UTC. See Figure \ref{fig:ray_tracing} for the definition of the tangential altitude)}
  \label{tab:r12_result}
  \begin{tblr}{
    colspec={X[c]X[c]X[c]X[c]X[c]X[c]},
    width=\textwidth,
    row{even} = {white,font=\small},
    row{odd}  = {bg=black!10,font=\small},
    row{1}    = {bg=black!20,font=\bfseries\small},
    row{2}    = {bg=black!20,font=\bfseries\small},
    hline{Z}  = {1pt,solid,black!60},
    rowsep=3pt
  }
\SetCell[r=2]{c} Tangential Altitude [km] &
\SetCell[c=5]{c} Mean Total Group Delay (99th-percentile) \\
$h_{\perp}$ &
R12 = 10.0 & R12 = 50.0 & R12 = 100.0 & R12 = 150.0 & R12 = 200.0 \\
0--500      & 12.05 (31.72) & 23.06 (49.55) & 39.97 (80.59) & 57.49 (107.89) & 75.01 (129.01) \\
500--1000   &  2.90 (7.90)  &  4.80 (13.71) &  8.20 (24.13) & 12.79 (37.69)  & 21.35 (82.27)  \\
1000--2000  &  2.21 (5.58)  &  2.91 (7.50)  &  4.04 (10.84) &  5.54 (17.01)  &  8.70 (35.88)  \\
2000--3000  &  1.50 (3.21)  &  1.84 (3.86)  &  2.37 (5.07)  &  3.09 (7.16)   &  4.48 (16.17)  \\
3000--4000  &  1.23 (2.85)  &  1.45 (3.27)  &  1.77 (3.90)  &  2.22 (4.64)   &  2.98 (6.54)   \\
4000--6000  &  0.84 (2.18)  &  0.96 (2.48)  &  1.16 (2.92)  &  1.46 (3.78)   &  2.50 (11.41)  \\
6000--8000  &  0.35 (1.14)  &  0.39 (1.27)  &  0.47 (1.45)  &  0.60 (1.69)   &  1.24 (8.18)   \\
8000--10000 &  0.19 (0.70)  &  0.21 (0.77)  &  0.25 (0.85)  &  0.31 (0.98)   &  0.74 (6.19)   \\
10000--15000&  0.06 (0.40)  &  0.06 (0.41)  &  0.08 (0.45)  &  0.10 (0.49)   &  0.32 (3.86)   \\
15000--20000&  0.01 (0.08)  &  0.02 (0.08)  &  0.02 (0.08)  &  0.03 (0.11)   &  0.13 (0.77)   \\
  \end{tblr}
\end{table}

%% file: sections/3_3_frequency.tex
\subsubsection{Simulation Results for Different Frequencies}
\label{sec:frequency}
\begin{figure}[ht!]
    \centering
    \includegraphics[width=\textwidth]{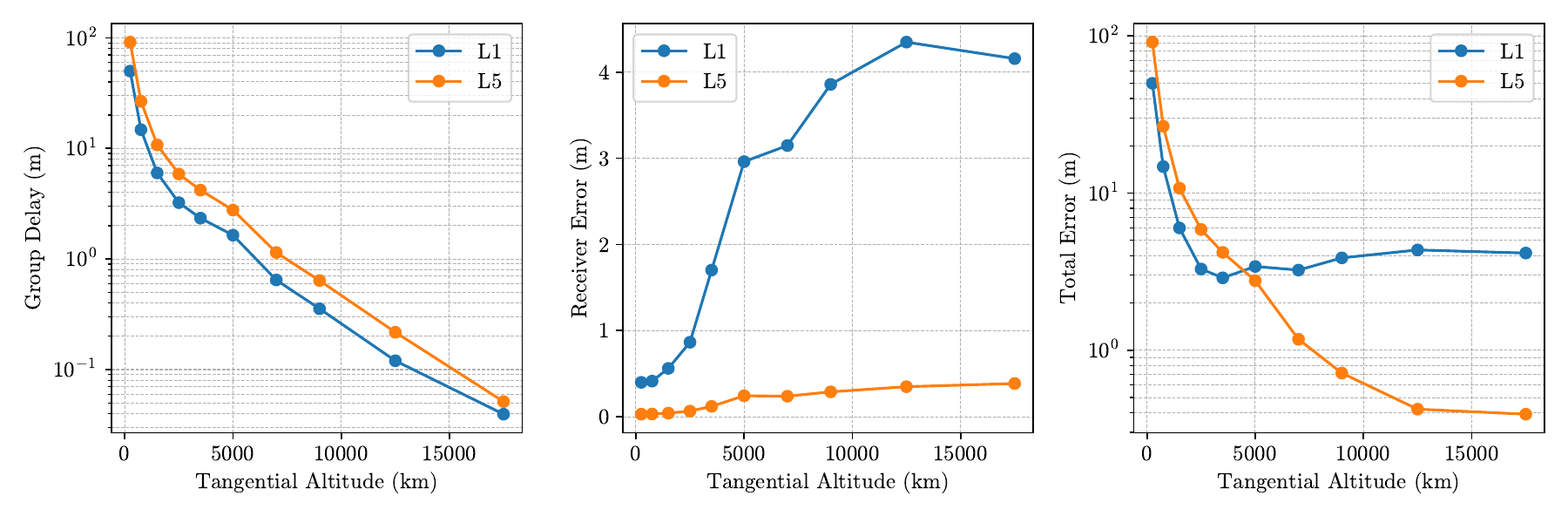}
    \caption{The total ionospheric and plasmaspheric group delay, the receiver noise, and the total UERE (=receiver noise + ionospheric delay) for GPS L1 and L5 signals with different tangential altitude ranges. The results are for the epoch 2025/01/01 12:00:00 UTC, $K_p$ index = 3.0, and $R12$ index = 167.24. The total UERE for the L1 is smaller than L5 for altitudes below 5000 km, due to the larger ionospheric and plasmaspheric delay added to the L5 signal. }
    \label{fig:multi_freq}
\end{figure}

We evaluated the effect of signal frequency by simulating the group delays and receiver thermal noise for the two civilian GPS signals: L1 (1575.42 MHz) and L5 (1176.45 MHz). The results are summarized in Table~\ref{tab:multi_freq} and Figure~\ref{fig:multi_freq}.

The simulations show that signal preference depends strongly on the tangential altitude of the ray path.
For rays with tangential altitudes below approximately 2000 km, the ionospheric and plasmaspheric group delays dominate the error budget, exceeding receiver noise by one to three orders of magnitude.
Across all altitudes, the mean group delays of the L1 signal are about 53–55 $\%$ of those of L5, whereas the L5 receiver noise is only 7–8 $\%$ of the L1 noise.
This creates a clear tradeoff: L1 offers smaller bias from group delay, while L5 provides lower variance from thermal noise.

To quantify the combined effect, we computed the total User Equivalent Range Error (UERE), defined as the mean absolute sum of the group delay and pseudorange noise over 100 samples,
\begin{equation}
\text{Total UERE} = \frac{1}{N_s} \sum_{i=1}^{N_s} \bigl| d_{\mathrm{total}} + \epsilon_{\rho} \bigr|,
\end{equation}
where $N_s=100$ and $\epsilon_{\rho}\sim \mathcal{N}(0,\sigma_{\rho})$ represents DLL thermal noise.
The noise variance $\sigma_{\rho}^2$ is derived from the carrier-to-noise ratio ($C/N_0$) using the expressions in~\citep{kaplan2005understanding}:
\begin{equation}
\sigma_{\rho}^2 =
\begin{cases}
\displaystyle \frac{B_{\mathrm{DLL}}}{2(C/N_0)}, d \left(1+\frac{2}{T(C/N_0)(2-d)}\right), & d \ge \dfrac{\pi}{T_{\mathrm{c}} B_{\mathrm{fe}}} \\
\displaystyle \frac{B_{\mathrm{DLL}}}{2(C/N_0)} \left(\frac{1}{T_{\mathrm{c}} B_{\mathrm{fe}}}\right) \left(1+\frac{1}{T(C/N_0)}\right), & d \le \dfrac{\pi}{T_{\mathrm{c}} B_{\mathrm{fe}}} \\
\displaystyle \frac{B_{\mathrm{DLL}}}{2(C/N_0)} \left(\frac{1}{T_{\mathrm{c}} B_{\mathrm{fe}}}+\frac{B_{\mathrm{fe}} T_{\mathrm{c}}}{\pi-1}\left(d-\frac{1}{T_{\mathrm{c}} B_{\mathrm{fe}}}\right)^2\right) \left(1+\frac{2}{T(C/N_0)(2-d)}\right), & \dfrac{1}{T_{\mathrm{c}} B_{\mathrm{fe}}}<d<\dfrac{\pi}{T_{\mathrm{c}} B_{\mathrm{fe}}}.
\end{cases}
\end{equation}
Parameter definitions and values are listed in Table~\ref{tab:dll_params}.

As seen in Table~\ref{tab:multi_freq} and Figure~\ref{fig:multi_freq}, L1 achieves a smaller total UERE than L5 for tangential altitudes up to roughly 5000 km because its lower group delay outweighs its higher receiver noise.
Conversely, L5 becomes advantageous for sidelobe paths where group delays are intrinsically small and thermal noise dominates.
Future work will examine optimal combinations of iono-free dual-frequency measurements and single-frequency tracking to minimize total error across different altitude regimes.

\begin{table}[htb]
  \caption{Parameters for Direct Lock Loop (DLL) Thermal Noise Calculation}
  \label{tab:dll_params}
  \begin{tblr}{
    colspec={X[l]X[c]Q[l,6.5cm]},
    width=\textwidth,
    row{even} = {white,font=\small},
    row{odd}  = {bg=black!10,font=\small},
    row{1}    = {bg=black!20,font=\bfseries\small},
    hline{Z}  = {1pt,solid,black!60},
    rowsep=3pt
  }
Parameter & Symbol & Value \\
Code tracking loop bandwidth & $B_{\mathrm{DLL}}$ & 0.1 Hz \\
Early--late correlator spacing & $d$ & 0.3 chips \\
Coherent integration time & $T$ & 20 ms \\
Chipping period & $T_{\mathrm{c}}$ & L1: 0.978 $\mu$s \quad L5: 0.0978 $\mu$s \\
Double-sided front-end bandwidth & $B_{\mathrm{fe}}$ & L1: 2.046 MHz \quad L5: 20.46 MHz \\
  \end{tblr}
\end{table}

%%%%%%%%%%%%%%%%%%%%%%%%%%%%%
% Table
\begin{table}[htb]
  \caption{Ionospheric and Plasmaspheric Group Delays for L1 and L5 signals (The error statistics for the results are for the epoch 2025/01/01 12:00:00 UTC, R12 = 167.24, $K_p$ index = 3.0)}
  \label{tab:multi_freq}
  \begin{tblr}{
    % Left column a bit wider, then a narrow "Frequency" col,
    % then three 3-col metric groups with equal narrow widths.
    colspec={Q[l, 2cm]Q[c, 1.5 cm]X[c]X[c]X[c]X[c]X[c]X[c]X[c]X[c]X[c]},
    width=\textwidth,
    % Header styling
    row{even} = {white,font=\small},
    row{odd}  = {bg=black!10,font=\small},
    row{1}    = {bg=black!20,font=\bfseries\small},
    row{2}    = {bg=black!20,font=\bfseries\small},
    hline{Z}  = {1pt,solid,black!60},
    rowsep=3pt
  }
% ---------------- Header row 1 ----------------
\SetCell[r=2]{l}{Tangential Altitude [km]} &
\SetCell[r=2]{l}{Frequency} &
\SetCell[c=3]{c}{Group Delay [m]} & & &
\SetCell[c=3]{c}{Receiver Noise [m]} & & & 
\SetCell[c=3]{c}{Total UERE [m]} \\
% -------- Header row 2 --------
& & Mean & 95\% & 99\% & Mean & 95\% & 99\% & Mean & 95\% & 99\% \\
% -------- Data --------
\SetCell[r=2]{l} 0--500      & L1 & 50.04 & 88.65 & 104.38 & 0.40 & 0.96 & 1.38 & 50.06 & 88.80 & 104.69 \\
                            & L5 & 91.43 & 161.32 & 190.66 & 0.03 & 0.07 & 0.10 & 91.43 & 161.33 & 190.67 \\
\SetCell[r=2]{l} 500--1000  & L1 & 14.76 & 45.76 & 62.51  & 0.43 & 1.05 & 1.41 & 14.77 & 45.77 & 62.74 \\
                            & L5 & 26.66 & 83.79 & 114.56 & 0.03 & 0.08 & 0.10 & 26.66 & 83.79 & 114.58 \\
\SetCell[r=2]{l} 1000--2000 & L1 & 6.00  & 13.19 & 23.04  & 0.56 & 1.43 & 1.97 & 6.00  & 13.48 & 22.69 \\
                            & L5 & 10.76 & 23.67 & 41.35  & 0.04 & 0.10 & 0.14 & 10.76 & 23.69 & 41.33 \\
\SetCell[r=2]{l} 2000--3000 & L1 & 3.23  & 6.66  & 9.09   & 0.89 & 2.27 & 3.14 & 3.32  & 7.50  & 9.38 \\
                            & L5 & 5.87  & 11.95 & 16.29  & 0.07 & 0.17 & 0.23 & 5.87  & 12.00 & 16.32 \\
\SetCell[r=2]{l} 3000--4000 & L1 & 2.34  & 4.73  & 5.02   & 1.66 & 4.34 & 6.98 & 2.87  & 6.69  & 8.92 \\
                            & L5 & 4.20  & 8.48  & 9.01   & 0.12 & 0.32 & 0.50 & 4.20  & 8.47  & 9.04 \\
\SetCell[r=2]{l} 4000--6000 & L1 & 1.64  & 3.81  & 5.71   & 2.94 & 8.05 & 12.01& 3.42  & 8.87  & 12.40 \\
                            & L5 & 2.77  & 6.51  & 10.24  & 0.25 & 0.72 & 1.14 & 2.77  & 6.59  & 10.33 \\
\SetCell[r=2]{l} 6000--8000 & L1 & 0.65  & 1.68  & 2.23   & 3.16 & 8.78 & 12.75& 3.24  & 8.90  & 12.83 \\
                            & L5 & 1.14  & 3.02  & 4.00   & 0.24 & 0.67 & 1.02 & 1.17  & 3.12  & 3.99 \\
\SetCell[r=2]{l} 8000--10000& L1 & 0.35  & 0.93  & 1.38   & 3.87 & 10.11& 14.15& 3.89  & 10.10 & 14.05 \\
                            & L5 & 0.64  & 1.67  & 2.47   & 0.29 & 0.76 & 1.14 & 0.72  & 1.86  & 2.45 \\
\SetCell[r=2]{l} 10000--15000& L1& 0.12  & 0.40  & 0.52   & 4.37 & 11.27& 15.23& 4.37  & 11.29 & 15.24 \\
                            & L5 & 0.22  & 0.68  & 0.94   & 0.35 & 0.92 & 1.33 & 0.42  & 1.10  & 1.52 \\
\SetCell[r=2]{l} 15000--20000& L1& 0.04  & 0.10  & 0.17   & 4.18 & 11.01& 15.45& 4.18  & 11.01 & 15.41 \\
                            & L5 & 0.05  & 0.15  & 0.29   & 0.39 & 1.05 & 1.50 & 0.39  & 1.05  & 1.51 \\
  \end{tblr}
\end{table}

%% file: sections/4_conclusion.tex
\section{Conclusion}
\label{sec:conclusion}
This study presented a detailed simulation framework for evaluating ionospheric and plasmaspheric delays affecting terrestrial GNSS signals received at the Moon. Using the GCPM and a custom ray-tracing algorithm, we quantified first-order, higher-order, and bending delays for both orbital and surface users across a wide range of geomagnetic and solar activity levels. 

The results show that mean group delays are generally near one meter but can reach beyond one hundred meters for low-altitude ray paths, particularly during periods of high solar activity.
Delays increase with increasing R12 index and decrease with higher geomagnetic $K_p$ index, reflecting the contraction of the plasmapause and the corresponding changes in electron density. 
Signal frequency further influences the trade-off between bias and variance: L1 signals experience smaller group delays but higher receiver noise, whereas L5 signals exhibit the opposite behavior, leading to altitude-dependent performance differences. 

Future work will focus on validating these simulations with upcoming LuGRE measurements, exploring mitigation techniques such as iono-free combinations and single-frequency optimization, and integrating delay estimation directly into navigation filters for lunar satellites. 
By clarifying the magnitude and variability of plasmaspheric propagation effects, this work establishes a foundation for reliable terrestrial GNSS-based positioning and timing in cislunar space.

%% file: references.bib
@inproceedings{parker2022lugre,
  author    = {Joel J.K. Parker and Fabio Dovis and Benjamin Anderson and Luigi Ansalone and Benjamin Ashman and Frank H. Bauer and Giuseppe D'Amore and Claudia Facchinetti and Samuele Fantinato and Gabriele Impresario and Stephen A. McKim and Efer Miotti and James J. Miller and Mario Musmeci and Oscar Pozzobon and Lauren Schlenker and Alberto Tuozzi and Lisa Valencia},
  title     = {The Lunar {GNSS} Receiver Experiment ({LuGRE})},
  booktitle = {Proceedings of the 2022 International Technical Meeting of The Institute of Navigation},
  year      = {2022},
  month     = {January},
  pages     = {420--437},
  address   = {Long Beach, California},
  publisher = {The Institute of Navigation},
  doi       = {10.33012/2022.18199},
}

@article{insidegnss2025lugre,
  author = {{Inside GNSS}},
  title  = {{LuGRE} Successfully Tracks {GNSS} Signals in Lunar Orbit},
  year   = {2025},
  month  = {February 24},
  url    = {https://insidegnss.com/lugre-successfully-tracks-gnss-signals-in-lunar-orbit/}
}

@inproceedings{giordano2022pathfinder,
  title     = {The {Lunar Pathfinder PNT} Experiment and {Moonlight} Navigation Service: The Future of Lunar Position, Navigation and Timing},
  author    = {Giordano, Pietro and Malman, Floor and Swinden, Richard and Zoccarato, Paolo and Ventura-Traveset, Javier,},
  booktitle = {Proceedings of the 2022 International Technical Meeting of The Institute of Navigation},
  year      = {2022},
  month     = {January},
  pages     = {632-642},
  address   = {Long Beach, California},
  publisher = {The Institute of Navigation},
  doi       = {10.33012/2022.18225},
  note      = {Article ID 18225. Accessed: March 01, 2025}
}

@inproceedings{Delepaut2024isl,
  author    = {Delépaut, Anaïs and Minetto, Alex and Giordano, Pietro and Dovis, Fabio},
  booktitle = {2024 International Conference on Localization and GNSS (ICL-GNSS)},
  title     = {{GNSS} Inter-User Ranging for Lunar Orbiters Flying the Spaceborne {Navimoon} Receiver},
  year      = {2024},
  volume    = {},
  number    = {},
  pages     = {1-7},
  doi       = {10.1109/ICL-GNSS60721.2024.10578517}
}

@inproceedings{Cheung2024PNT,
  author    = {Cheung, Kar-Ming and Jun, William W. and Bhamidipati, Sriramya and Carter, Paul},
  booktitle = {2024 IEEE Aerospace Conference},
  title     = {Ground-Assisted Position Navigation and Timing ({PNT}) for {Moon} and {Mars}},
  year      = {2024},
  volume    = {},
  number    = {},
  pages     = {1-19},
  doi       = {10.1109/AERO58975.2024.10520998}
}

@article{jakowski2018,
  title     = {{NPSM}: A New Empirical Model for the Plasmasphere},
  author    = {Jakowski, N. and Hoque, M.},
  journal   = {Advances in Space Research},
  volume    = {62},
  number    = {3},
  pages     = {792--800},
  year      = {2018},
  publisher = {Elsevier}
}

@article{Gallagher2000GCPM,
  author   = {Gallagher, D. L. and Craven, P. D. and Comfort, R. H.},
  title    = {Global core plasma model},
  journal  = {Journal of Geophysical Research: Space Physics},
  volume   = {105},
  number   = {A8},
  pages    = {18819-18833},
  doi      = {https://doi.org/10.1029/1999JA000241},
  url      = {https://agupubs.onlinelibrary.wiley.com/doi/abs/10.1029/1999JA000241},
  eprint   = {https://agupubs.onlinelibrary.wiley.com/doi/pdf/10.1029/1999JA000241},
  year     = {2000}
}

@article{Antonio2012,
  author  = {Angrisano, Antonio and Gaglione, Salvatore and Gioia, Ciro and Massaro, Marco and Robustelli, Umberto},
  year    = {2012},
  month   = {01},
  pages   = {1-20},
  title   = {Assessment of {NeQuick} Ionospheric model for {Galileo} single-frequency users},
  journal = {Acta Geophysica},
  doi     = {10.2478/s11600-013-0116-2}
}

@article{Jakowski2018NSPM,
  author  = {Jakowski, Norbert and Hoque, Mohammed Mainul},
  title   = {A new electron density model of the plasmasphere for operational applications and services},
  doi     = {10.1051/swsc/2018002},
  url     = {https://doi.org/10.1051/swsc/2018002},
  journal = {J. Space Weather Space Clim.},
  year    = 2018,
  volume  = 8,
  pages   = {A16}
}

@inproceedings{Umut2017IRIPlas,
  author    = {Sezen, Umut and Gulyaeva, Tamara L. and Arikan, Feza},
  booktitle = {2017 XXXIInd General Assembly and Scientific Symposium of the International Union of Radio Science (URSI GASS)},
  title     = {Online international reference ionosphere extended to plasmasphere ({IRI-Plas}) model},
  year      = {2017},
  volume    = {},
  number    = {},
  pages     = {1-4},
  doi       = {10.23919/URSIGASS.2017.8105426}
}

@article{SHI2012,
  title    = {An improved approach to model ionospheric delays for single-frequency Precise Point Positioning},
  journal  = {Advances in Space Research},
  volume   = {49},
  number   = {12},
  pages    = {1698-1708},
  year     = {2012},
  issn     = {0273-1177},
  doi      = {https://doi.org/10.1016/j.asr.2012.03.016},
  url      = {https://www.sciencedirect.com/science/article/pii/S0273117712001822},
  author   = {Chuang Shi and Shengfeng Gu and Yidong Lou and Maorong Ge}
}

@article{Iiyama2023Diff,
  title   = {Positioning and Timing of Distributed Lunar Satellites via Terrestrial {GPS} Differential Carrier Phase Measurements},
  author  = {Iiyama, Keidai and Gao, Grace},
  journal = {Proceedings of the 36th International Technical Meeting of the Satellite Division of The Institute of Navigation (ION GNSS+ 2023)},
  pages   = {1511-1529},
  year    = {2023},
  doi     = {10.33012/2023.19385}
}

@article{Peng2017,
  author   = {Chen, Peng and Yao, Yibin and Li, Qinzheng and Yao, Wanqiang},
  title    = {Modeling the plasmasphere based on {LEO} satellites onboard {GPS} measurements},
  journal  = {Journal of Geophysical Research: Space Physics},
  volume   = {122},
  number   = {1},
  pages    = {1221-1233},
  doi      = {https://doi.org/10.1002/2016JA023375},
  url      = {https://agupubs.onlinelibrary.wiley.com/doi/abs/10.1002/2016JA023375},
  eprint   = {https://agupubs.onlinelibrary.wiley.com/doi/pdf/10.1002/2016JA023375},
  year     = {2017}
}

@article{Matsumoto2023,
  author  = {Matsumoto, Takehiro and Sakamoto, Takushi and Nakajima, Ayano and Hamada, Kiyoshi and Nakamura, Shinichi},
  title   = {Plasmaspheric Correction with Global Core Plasma Model ({GCPM}) for {GPS}-Based {GEO} Precise Orbit Determination},
  journal = {Proceedings of the 36th International Technical Meeting of the Satellite Division of The Institute of Navigation (ION GNSS+ 2023)},
  year    = {2023},
  page    = {1487-1498},
  doi     = {10.33012/2023.19372}
}

@article{Hoque2019,
  author  = {Hoque, M. Mainul and Jakowski, Norbert and Osechas, Okuary and Berdermann, Jens,},
  title   = {Fast and Improved Ionospheric Correction for {Galileo} Mass Market Receivers},
  journal = {Proceedings of the 32nd International Technical Meeting of the Satellite Division of The Institute of Navigation (ION GNSS+ 2019)},
  year    = {2019},
  page    = {3377-3389},
  doi     = {10.33012/2019.17106}
}

@article{Klobuchar1987,
  author   = {Klobuchar, John A.},
  journal  = {IEEE Transactions on Aerospace and Electronic Systems},
  title    = {Ionospheric Time-Delay Algorithm for Single-Frequency {GPS} Users},
  year     = {1987},
  volume   = {AES-23},
  number   = {3},
  pages    = {325-331},
  doi      = {10.1109/TAES.1987.310829}
}

@article{Vila2025fusion,
  author={Vila, Guillem Casadesus and Gao, Grace},
  journal={2025 IEEE Aerospace Conference}, 
  title={Sensor Fusion for Autonomous Orbit Determination and Time Synchronization in Lunar Orbit}, 
  year={2025},
  volume={},
  number={},
  pages={1-12},
  doi={10.1109/AERO63441.2025.11068682}}

@INPROCEEDINGS{Vila2025LuPNT,
  author={Vila, Guillem Casadesus and Iiyama, Keidai and Gao, Grace},
  booktitle={2025 IEEE Aerospace Conference}, 
  title={{LuPNT:} An Open-Source Simulator for Lunar Communications, Positioning, Navigation, and Timing}, 
  year={2025},
  volume={},
  number={},
  pages={1-18},
  doi={10.1109/AERO63441.2025.11068501}}

@article{KLIMENKO2015,
  title    = {The global morphology of the plasmaspheric electron content during Northern winter 2009 based on {GPS/COSMIC} observation and {GSM TIP} model results},
  journal  = {Advances in Space Research},
  volume   = {55},
  number   = {8},
  pages    = {2077-2085},
  year     = {2015},
  note     = {INTERNATIONAL REFERENCE IONOSPHERE AND GLOBAL NAVIGATION SATELLITE SYSTEMS},
  issn     = {0273-1177},
  doi      = {https://doi.org/10.1016/j.asr.2014.06.027},
  url      = {https://www.sciencedirect.com/science/article/pii/S0273117714003895},
  author   = {M.V. Klimenko and V.V. Klimenko and I.E. Zakharenkova and Iu.V. Cherniak}
}

@article{Capuano2017,
  author  = {Capuano, Vincenzo and Blunt, Paul and Botteron, Cyril and Farine, Pierre Andr{\'{e}}},
  doi     = {10.1002/navi.185},
  issn    = {00281522},
  journal = {Navigation, Journal of the Institute of Navigation},
  number  = {3},
  pages   = {323--338},
  title   = {{Orbital Filter Aiding of a High Sensitivity {GPS} Receiver for Lunar Missions}},
  volume  = {64},
  year    = {2017}
}

@article{Menzione2024,
  author    = {Menzione, F and Sgammini, M and Paonni, M},
  title     = {Reconstruction of {Galileo} Constellation Antenna Pattern for space service volume applications},
  journal   = {Publications Office of the European Union},
  year      = {2024},
  doi       = {doi/10.2760/765842}
}

@article{Fischer2022,
  author = {Aaron Fischer},
  title  = {{GPS III} {Earth} Coverage ({EC)} Antenna Patterns},
  year   = {2022}
}

@article{donaldson2020characterization,
  title   = {Characterization of on-orbit {GPS} transmit antenna patterns for space users},
  author  = {Donaldson, Jennifer E and Parker, Joel JK and Moreau, Michael C and Highsmith, Dolan E and Martzen, Philip D},
  journal = {NAVIGATION, Journal of the Institute of Navigation},
  volume  = {67},
  number  = {2},
  pages   = {411--438},
  year    = {2020},
  doi     = {https://doi.org/10.1002/navi.361}
}

@article{Marquis2015,
  author  = {Marquis, Willard A. and Reigh, Daniel L.},
  doi     = {https://doi.org/10.1002/navi.123},
  issn    = {00281522},
  journal = {Navigation, Journal of the Institute of Navigation},
  number  = {4},
  pages   = {329--347},
  title   = {The {GPS} Block {IIR} and {IIR-M} Broadcast {L}-band Antenna Panel: Its Pattern and Performance},
  volume  = {62},
  year    = {2015}
}

@article{NavimoonIOD,
  author = {{SpacePNT}},
  title  = {{Navimoon IOD}: {Earth Moon} {GNSS} Spaceborne Receiver for In-Orbit Demonstration},
  year   = {2024},
  url    = {https://navisp.esa.int/uploads/files/documents/NaviMIOD_Final_Presentation.pdf}
}

@book{kaplan2017understanding,
  title     = {Understanding {GPS}/{GNSS}: {Principles} and applications},
  author    = {Kaplan, Elliott D and Hegarty, Christopher},
  year      = {2017},
  publisher = {Artech House}
}

@article{Nelder1965Simplex,
  author  = {Nelder, J. A. and Mead, R.},
  title   = {A simplex method for function minimization},
  journal = {The Computer Journal},
  volume  = {7},
  number  = {4},
  pages   = {308--313},
  year    = {1965},
  doi     = {10.1093/comjnl/7.4.308}
}

@book{Born1980Optics,
  author    = {Born, Max and Wolf, Emil},
  title     = {Principles of Optics},
  edition   = {6},
  publisher = {Pergamon Press},
  year      = {1980},
  address   = {Oxford, United Kingdom},
}

@article{Caruso2023RayTracing,
  author  = {Caruso, A. and Bourgoin, A. and Togni, A. and Zannoni, M. and Tortora, P.},
  title   = {Radio Occultation Data Analysis with Analytical Ray-Tracing},
  journal = {Radio Science},
  volume  = {58},
  pages   = {e2023RS007740},
  year    = {2023},
  doi     = {10.1029/2023RS007740}
}

@article{Hoque2011BendingCorrection,
  author  = {Hoque, M. M. and Jakowski, N.},
  title   = {Ionospheric Bending Correction for {GNSS} Radio Occultation Signals},
  journal = {Radio Science},
  volume  = {46},
  pages   = {RS0D06},
  year    = {2011},
  doi     = {10.1029/2010RS004583}
}

@article{Bilitza2024IRI,
  author  = {Bilitza, D. and Truhlik, V. and Yoshihara, O. and Moldwin, M. B.},
  title   = {Development and Improvement of the {International Reference Ionosphere} with Special Emphasis on the Topside and Extension to the Plasmasphere},
  journal = {Annals of Geophysics},
  volume  = {67},
  number  = {4},
  pages   = {SA443},
  year    = {2024},
  doi     = {10.4401/ag-9145}
}

@article{Strangeways2000HomingIn,
  author  = {Strangeways, Hal J.},
  title   = {Effect of Horizontal Gradients on Ionospherically Reflected or Transionospheric Paths Using a Precise Homing-in Method},
  journal = {Journal of Atmospheric and Solar-Terrestrial Physics},
  volume  = {62},
  number  = {15},
  pages   = {1361--1376},
  year    = {2000},
  issn    = {1364-6826},
  doi     = {10.1016/S1364-6826(00)00150-4},
  url     = {https://www.sciencedirect.com/science/article/pii/S1364682600001504}
}

@techreport{LCRNS2025,
  author      = {Grant Ryden and Michael Volle},
  title       = {{NASA} Lunar Communications Relay and Navigation Systems
                 ({LCRNS}) Reference Constellation 3.1},
  institution = {NASA Goddard Space Flight Center},
  year        = {2025},
  note        = {Available at \url{https://esc.gsfc.nasa.gov/static-files/LCRNS_Reference_Constellation_White_Paper_03_2025.pdf}}
}

@article{Hoque2008Estimate,
  author   = {Hoque, M. Mainul and Jakowski, N.},
  journal  = {Radio Science},
  title    = {Estimate of higher order ionospheric errors in {GNSS} positioning},
  year     = {2008},
  volume   = {43},
  number   = {05},
  pages    = {1-15},
  doi      = {10.1029/2007RS003817}
}

@article{pharlap,
  author = {Manuel Cervera},
  title  = {{PHaRLAP} - Provision of High-frequency Raytracing Laboratory for Propagation studies},
  year   = {2025},
  url    = {https://www.dst.defence.gov.au/our-technologies/pharlap-provision-high-frequency-raytracing-laboratory-propagation-studies}
}

@article{Delepaut2020,
  title = {Use of {{GNSS}} for Lunar Missions and Plans for Lunar In-Orbit Development},
  author = {Del{\'e}paut, Ana{\"i}s and Giordano, Pietro and {Ventura-Traveset}, Javier and Blonski, Daniel and Sch{\"o}nfeldt, Miriam and Schoonejans, Philippe and Aziz, Sarmad and Walker, Roger},
  year = {2020},
  month = dec,
  journal = {Advances in Space Research},
  series = {Scientific and {{Fundamental Aspects}} of {{GNSS}} - {{Part}} 1},
  volume = {66},
  number = {12},
  pages = {2739--2756},
  issn = {0273-1177},
  doi = {https://doi.org/10.1016/j.asr.2020.05.018},
}

@article{Iiyama_LuPNT,
    author = {Keidai Iiyama and Guillem Casadesus Vila and Grace Gao},
    title = {{LuPNT}: Open-Source Simulator for Lunar Positioning, Navigation, and Timing},
    journal={Proceedings of the 36th International Technical Meeting of the Satellite Division of The Institute of Navigation (ION GNSS+ 2023)},
    year={2023},
    pages={1499--1510},
}

@article{Iiyama2024TDCPNavigation,
  author       = {Iiyama, Keidai and Bhamidipati, Sriramya and Gao, Grace},
  title        = {Precise Positioning and Timekeeping in a Lunar Orbit via Terrestrial {GPS} Time-Differenced Carrier-Phase Measurements},
  volume       = {71},
  number       = {1},
  elocation-id = {navi.635},
  year         = {2024},
  doi          = {10.33012/navi.635},
  publisher    = {Institute of Navigation},
  issn         = {0028-1522},
  journal      = {NAVIGATION: Journal of the Institute of Navigation}
}

@article{iiyama2025plasma,
  title = {Plasmaspheric Delay Characterization and Comparison of Mitigation Methodologies for Lunar Terrestrial {GNSS} Receivers},
  pages = {},
  year = {2025},
  month = {September},
  journal = {Proceedings of the Institute of Navigation GNSS+ conference (ION GNSS+ 2025)},
  author = {Iiyama, Keidai and Gao, Grace},
}

@book{kaplan2005understanding,
  title     = {Understanding {GPS}: principles and applications},
  author    = {Kaplan, Elliott and Hegarty, Christopher},
  year      = {2005},
  publisher = {Artech house}
}

@incollection{Jakowski12,
author = {M. Mainul Hoque and Norbert Jakowski},
title = {Ionospheric Propagation Effects on GNSS Signals and New Correction Approaches},
booktitle = {Global Navigation Satellite Systems - Signal, Theory and Applications},
publisher = {IntechOpen},
address = {London},
year = {2012},
editor = {Shuanggen Jin},
chapter = {16},
doi = {10.5772/30090},
url = {https://doi.org/10.5772/30090}
}

@techreport{Jones1975RayTracing,
  author      = {Jones, R. M. and Stephenson, J. J.},
  title       = {A Versatile Three-Dimensional Ray Tracing Computer Program for Radio Waves in the Ionosphere},
  institution = {NASA STI/Recon Technical Report},
  number      = {76-25476},
  year        = {1975},
  note        = {NASA Technical Report}
}

@article{Coleman2011RayTracing,
  author  = {Coleman, C. J.},
  title   = {Point-to-Point Ionospheric Ray Tracing by a Direct Variational Method},
  journal = {Radio Science},
  volume  = {46},
  pages   = {RS5016},
  year    = {2011},
  doi     = {10.1029/2011RS004744}
}
